\documentclass[conference]{IEEEtran}

\newif\ifanonymous
\anonymousfalse

\usepackage{cite}
\usepackage{amsmath,amssymb,amsfonts}
\usepackage{algorithmic}
\usepackage{graphicx}
\usepackage{textcomp}
\usepackage{xcolor}
\def\BibTeX{{\rm B\kern-.05em{\sc i\kern-.025em b}\kern-.08em
    T\kern-.1667em\lower.7ex\hbox{E}\kern-.125emX}}

\usepackage{tikz}
\usepackage{graphicx}
\usepackage{multirow}
\usepackage{amsmath}
\usepackage{amssymb}
\usepackage{wrapfig}   
\usepackage{float}
\usepackage{algorithm}
\usepackage{mdframed}
\usepackage{tcolorbox}
\usepackage{subcaption}
\usepackage{listings}
\usepackage{lipsum}
\usepackage{array}
\usepackage{chngcntr}
\usepackage{ragged2e}
\tcbuselibrary{breakable}
\usepackage{booktabs}
\usepackage{makecell}
\usepackage{caption}
\usepackage{xcolor} 
\usepackage{pifont}  
\usepackage{enumitem}

\usepackage{booktabs}
\usepackage{multirow}
\usepackage[table]{xcolor}

\usepackage[table]{xcolor} 
\definecolor{lightgray}{gray}{0.9} 
\usepackage{tcolorbox}
\tcbuselibrary{skins}
\usepackage{booktabs}
\usepackage{colortbl}
\usepackage[table]{xcolor}
\usepackage{siunitx}
\usepackage{xfp} 
\newcommand{\best}[1]{\textbf{#1}}
\newcommand{\second}[1]{\underline{#1}}

\usepackage{xstring}   

\definecolor{myGreen}{RGB}{0, 150, 0}
\definecolor{myRed}{RGB}{200, 0, 0}
\newcommand{\perf}[2]{%
    #1%
    \rlap{$
        \,_{\IfBeginWith{#2}{-}%
            {\color{myGreen}\text{\tiny{(#2)}}}%
            {\color{myRed}\text{\tiny{(#2)}}}%
        }
    $}%
}

\newtcolorbox{definitionbox}[1][]{%
  colback=blue!5,       
  colframe=blue!50!black, 
  coltitle=white,       
  colbacktitle=blue!60!black, 
  boxrule=1.5pt,                 
  rounded corners,               
  fonttitle=\bfseries,  
  enhanced,
  attach boxed title to top left={yshift=-2mm,xshift=2mm},
  boxed title style={
    rounded corners,
    borderline west={0pt}{0pt}{white}, 
    borderline east={0pt}{0pt}{white},
    borderline north={0pt}{0pt}{white},
    borderline south={0pt}{0pt}{white},
  },
  title=Definition,
  #1
}

\definecolor{thm}{RGB}{69, 53, 193}

\newcounter{assump}[section]

\newtcolorbox{assumbox}[1][]{%
  colback=blue!5,       
  colframe=blue!50!black, 
  coltitle=white,       
  colbacktitle=blue!60!black, 
  boxrule=1.5pt,                 
  rounded corners,               
  fonttitle=\bfseries,  
  enhanced,
  breakable,
  attach boxed title to top left={yshift=-2mm,xshift=2mm},
  boxed title style={
    rounded corners,
    borderline west={0pt}{0pt}{white}, 
    borderline east={0pt}{0pt}{white},
    borderline north={0pt}{0pt}{white},
    borderline south={0pt}{0pt}{white},
  },
  before upper={\refstepcounter{assump}},
  #1
}

\newtcolorbox{thmbox}[1][]{%
  colback=green!5,       
  colframe=green!50!black, 
  coltitle=white,       
  colbacktitle=green!60!black, 
  boxrule=1.5pt,                 
  rounded corners,               
  fonttitle=\bfseries,  
  enhanced,
  breakable,
  attach boxed title to top left={yshift=-2mm,xshift=2mm},
  boxed title style={
    rounded corners,
    borderline west={0pt}{0pt}{white}, 
    borderline east={0pt}{0pt}{white},
    borderline north={0pt}{0pt}{white},
    borderline south={0pt}{0pt}{white},
  },
  #1
}

\definecolor{upcolor}{RGB}{56, 142, 60}   
\definecolor{downcolor}{RGB}{211, 47, 47} 
\definecolor{groupgray}{gray}{0.95}       
\definecolor{mybrown}{HTML}{E3D9CC}
\definecolor{mygreen}{HTML}{bcd1ca}
\definecolor{mydarkbrown}{HTML}{d97557}
\definecolor{mydarkgreen}{HTML}{3b8f7a}
\definecolor{myblue}{HTML}{6c9ecc}
\definecolor{mypink}{HTML}{c26584}
\definecolor{coral}{HTML}{EBAF9A}      
\definecolor{peach}{HTML}{F3C9B8}      
\definecolor{mustard}{HTML}{D6B04A}    
\definecolor{marigold}{HTML}{F1B86B}   
\definecolor{slate}{HTML}{92A0A8}      
\definecolor{sky}{HTML}{A7C9E8}        
\definecolor{teal}{HTML}{67B2A8}       
\definecolor{seafoam}{HTML}{B9E3D6}    
\definecolor{olive}{HTML}{A6B182}      
\definecolor{mint}{HTML}{C8EAD7}       
\definecolor{plum}{HTML}{8E5A7A}       
\definecolor{sienna}{HTML}{B87459}     
\definecolor{ochre}{HTML}{C7A35A}      
\definecolor{charcoal}{HTML}{5B6166}   
\definecolor{navy}{HTML}{324A6B}       
\definecolor{lavender}{HTML}{C9B3E0}

\newcommand{\cmark}{\ding{51}}
\newcommand{\xmark}{\ding{55}}

\newcommand{\Qwenemoji}{\includegraphics[height=1.1\fontcharht\font`\B]{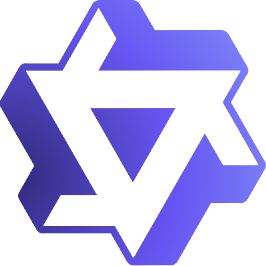}}

\newcommand{\Openaiemoji}{\includegraphics[height=1.1\fontcharht\font`\B]{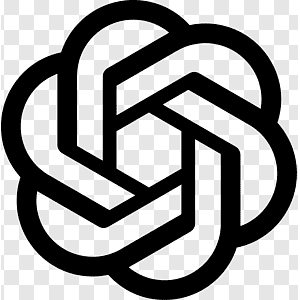}}
\newcommand{\claudemoji}{\includegraphics[height=1.1\fontcharht\font`\B]{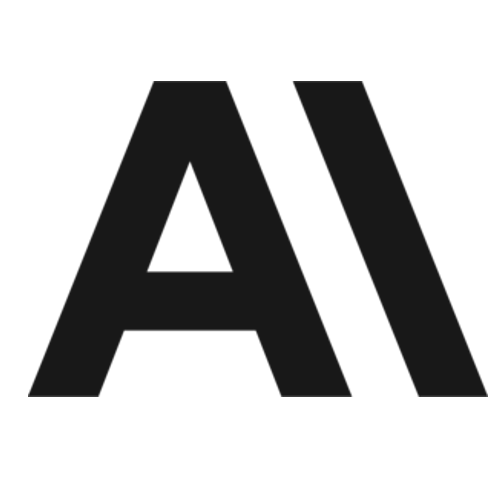}}

\newcommand{\deepseekemoji}{\includegraphics[height=1.1\fontcharht\font`\B]{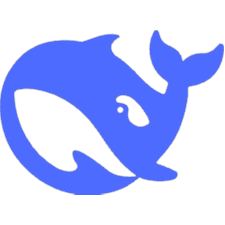}}

\definecolor{myPurple}{RGB}{106,12,173}
\definecolor{myGreen}{RGB}{34, 140, 35}
\usepackage{xspace}
\newcommand{\ours}{DataCOPE\xspace}

\begin{document}

\title{Unsupervised Skill Discovery for  Agentic Data Analysis}

\ifanonymous
\author{
\IEEEauthorblockN{Anonymous Authors}
}
\else
\author{
\IEEEauthorblockN{1\textsuperscript{st} Zhisong Qiu\textsuperscript{*}}
\IEEEauthorblockA{\textit{Zhejiang University} \\
Hangzhou, China \\
qiuzhisong@zju.edu.cn
}
\and
\IEEEauthorblockN{2\textsuperscript{nd} Kangqi Song\textsuperscript{*}}
\IEEEauthorblockA{
\textit{Zhejiang University} \\
Hangzhou, China \\
kangqi.song@outlook.com
}
\and
\IEEEauthorblockN{3\textsuperscript{rd} Shengwei Tang}
\IEEEauthorblockA{
\textit{Zhejiang University} \\
Hangzhou, China \\
tangshengwei40@gmail.com
}
\and
\IEEEauthorblockN{4\textsuperscript{th} Shuofei Qiao}
\IEEEauthorblockA{
\textit{Zhejiang University} \\
Hangzhou, China \\
shuofei@zju.edu.cn
}
\and
\IEEEauthorblockN{5\textsuperscript{th} Lei Liang}
\IEEEauthorblockA{
\textit{Ant Group} \\
Hangzhou, China \\
leywar.liang@antgroup.com
}
\and
\IEEEauthorblockN{6\textsuperscript{th} Huajun Chen}
\IEEEauthorblockA{\textit{Zhejiang University} \\
Hangzhou, China \\
huajunsir@zju.edu.cn
}
\and
\IEEEauthorblockN{7\textsuperscript{th} Shumin Deng\textsuperscript{$\dagger$}}
\IEEEauthorblockA{
\textit{Zhejiang University} \\
Hangzhou, China \\
231sm@zju.edu.cn}
}
\fi

\maketitle

\begingroup
\renewcommand{\thefootnote}{\fnsymbol{footnote}}
\setcounter{footnote}{0}
\footnotetext[1]{Equal contribution.}
\footnotetext[2]{Corresponding author.}
\endgroup
\setcounter{footnote}{0}


\begin{abstract}
Inference-time skill augmentation provides a lightweight way to improve data-analytic agents by injecting reusable procedural knowledge without updating model parameters. However, discovering effective skills for data analysis remains challenging, as reliable supervision is expensive and success criteria vary across analytical formats. This raises the key question of how to discover reusable data-analysis skills from unlabeled exploration alone. We propose \ours{}, an unsupervised verifier-guided skill discovery framework for data-analytic agents. \ours{} derives verifier signals from the exploration trajectories and uses them to characterize relative quality or aggreement among trajectories. It iteratively coordinates a Data-Analytic Agent for trajectory generation, an Unsupervised Verifier for signal extraction, and a Skill Manager for contrastive skill distillation. For report-style analysis, we instantiate the verifier as an Adaptive Checklist Verifier that derives task-specific criteria, scores reports by verifiable coverage, and iteratively refines the checklist. For reasoning-style analysis, we instantiate it as an Answer Agreement Verifier that groups trajectories by answer agreement and uses self-consistency as an auxiliary signal. We evaluate \ours{} on report-style analysis from Deep Data Research and reasoning-style analysis from DABStep. Across both settings, \ours{} consistently improves held-out performance over baselines. Averaged across four model settings, \ours{} improves the mean score by 9.71\% and 32.30\% on report-style and reasoning-style tasks respectively.
\end{abstract}


\section{Introduction}

\begin{figure}[t]
    \centering
    \includegraphics[width=\linewidth]{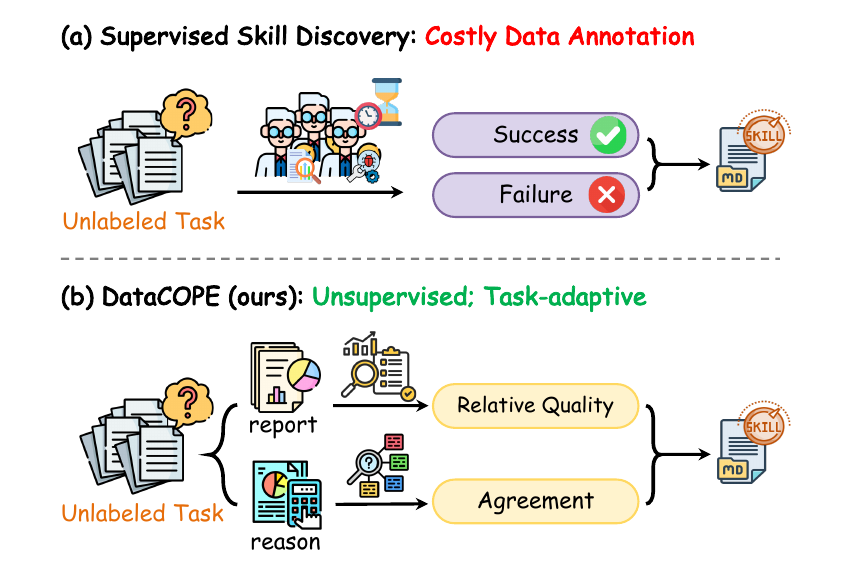}
    \vspace{-4mm}
    \caption{
    Supervised skill discovery requires costly data annotation.
    \ours{} instead performs unsupervised skill discovery by deriving task-adaptive verifier signals from unlabeled exploration trajectories and distilling them into reusable skills.
    }
    \vspace{-4mm}
    \label{fig:teaser}
\end{figure}



Automated data analysis has long been a central goal in data mining, aiming to transform raw data into reliable findings with minimal human intervention~\cite{dataagent_survey, dsagent_survey, dr_survey}. 
Recent LLM agents have made substantial progress toward this goal by automating complex analytical workflows, including data inspection, tool use, hypothesis exploration, and report generation~\cite{datainterpreter, agenticdata, dsstar, agentada, datawiseagent, datacopilot, dagent, datamind, deepanalyze}. 
Despite advances, data-analysis tasks vary widely in goals, data formats, and analytical requirements, making it difficult to rely on fixed pipelines across domains. 
Inference-time skill augmentation offers an alternative by injecting reusable procedural knowledge into agents, enabling them to adapt their exploration strategies, analytical choices, and error-avoidance behaviors without updating model parameters~\cite{anthropic2026skills, sok_skill, agent_skills}.

However, as shown in Figure~\ref{fig:teaser}, discovering such skills for data-analytic agents remains challenging. 
Existing skill synthesis and refinement methods typically rely on observable quality signals to identify useful behaviors and failure~\cite{skillx, trace2skill, evoskill, coevoskills, skillopt, skillclaw, skillos}.
These signals may come from successful demonstrations, failure cases, or human feedback.
In data-analysis scenarios, such signals are often unavailable or difficult to construct for two main reasons:
\emph{\textbf{(1) Reliable supervision requires high-effort analytical annotation.}}
Unlike tasks whose labels can be obtained by checking a short answer or applying a predefined criterion, data-analysis tasks require annotators to understand the task objective, inspect the associated data resources, and assess whether the analytical process and final output are well supported by the data.
\emph{\textbf{(2) Success criteria vary across analytical formats.}}
Beyond the cost of annotation, data-analysis tasks also differ in what constitutes a valid quality signal.
For reasoning-oriented tasks, success is often judged by whether the derived final answer is consistent with an expected solution.
By contrast, open-ended analytical tasks usually lack a unique target answer and are instead judged by report completeness, evidence-supported claims, and analytical insight.
Such heterogeneity makes it difficult to define a single signal that can reliably compare unlabeled trajectories across different forms of data analysis.

To address these challenges, we propose \textbf{\ours{}}, an \emph{unsupervised verifier-guided skill discovery} framework for data-analytic agents. 
Instead of relying on external supervision, \ours{} derives verifier signals from the agent's own exploration trajectories. 
These signals do not directly certify trajectory correctness, but capture relative quality or agreement among trajectories, thereby providing the contrastive evidence needed for skill discovery. 
Specifically, \ours{} iteratively coordinates a \emph{Data-Analytic Agent} that samples exploration trajectories, an \emph{Unsupervised Verifier} that extracts task-dependent signals and organizes trajectories into contrastive groups, and a \emph{Skill Manager} that distills reusable analytical procedures from these groups. 
We instantiate the verifier for two representative settings. For open-ended report-style tasks, an Adaptive Checklist Verifier generates task-specific criteria and scores reports by verifiable coverage, with iterative refinement to reduce checklist incompleteness; for fixed-answer reasoning-style tasks, an Answer Agreement Verifier groups trajectories by final answers and uses self-consistency as an auxiliary uncertainty signal.
Through iterative trajectory generation, unsupervised verification, and contrastive skill distillation, \ours{} discovers reusable skills that transfer to held-out data-analysis tasks without ground-truth answers, success labels, or human annotations.

We evaluate \ours{} on two categories of data-analysis benchmarks: report-style analysis from Deep Data Research~\cite{ddr} and reasoning-style analysis from DABStep~\cite{dabstep}.
Across both settings, \ours{} consistently improves held-out performance over strong baselines.
Averaged across four matched base models, \ours{} yields substantial gains on both report-style tasks and reason-style tasks, improving the mean score by 9.71\% and 32.30\% respectively.
We further conduct comprehensive analyses on iterative refinement, verifier-component ablations, skill granularity, data-analytic agent ablations, label efficiency, and inference cost. Our main contributions are summarized as follows:
\begin{itemize}[leftmargin=*]
    \item We propose \ours{}, an unsupervised verifier-guided skill discovery framework that improves data-analytic agents by iteratively coordinating trajectory generation, unsupervised verification, and contrastive skill distillation without using ground-truth answers or success labels.

    \item We design unsupervised verifiers for two representative data-analysis settings: an Adaptive Checklist Verifier with checklist refinement for report-style tasks, and an Answer Agreement Verifier with self-consistency estimation for reasoning-style tasks.

    \item We provide systematic empirical evidence that verifier-derived unsupervised signals are essential for skill discovery, demonstrating that the discovered skills improve held-out generalization and transfer across different models.
\end{itemize}
\section{Preliminary}
\label{sec:preliminary}

\subsection{Data-Analytic Agents}


Given a data analysis task space $\mathcal{U}$, where each task $u \in \mathcal{U}$ consists of a user query and associated data resources, we formulate the interaction process for each task $u$ as a task-conditioned POMDP:
\begin{equation}
\mathcal{M}_u = \langle \mathcal{X}, \mathcal{A}, T, \mathcal{O}, \Omega, R^\star \rangle .
\end{equation}
Here, $\mathcal{X}$ is the underlying environment state space, encompassing the state of the code interpreter, data files (e.g., csv files or databases), intermediate variables, and other relevant context. $\mathcal{A}$ denotes the action space, such as Python/SQL code generation and final-answer submission. The transition function $T$ governs the environment state update after executing an action, i.e., $x_{t+1}=T(x_t,a_t)$. Due to partial observability, the agent cannot directly access $x_t$. It receives an observation $o_t \in \mathcal{O}$ governed by the observation function $\Omega$. The hidden reward function $R^\star(x_T,u)\in[0,1]$ evaluates whether the terminal state $x_T$ satisfies the task requirements. Notably, this reward is unavailable during the agent's execution.


\begin{figure*}[t]
    \centering
    \includegraphics[width=\linewidth]{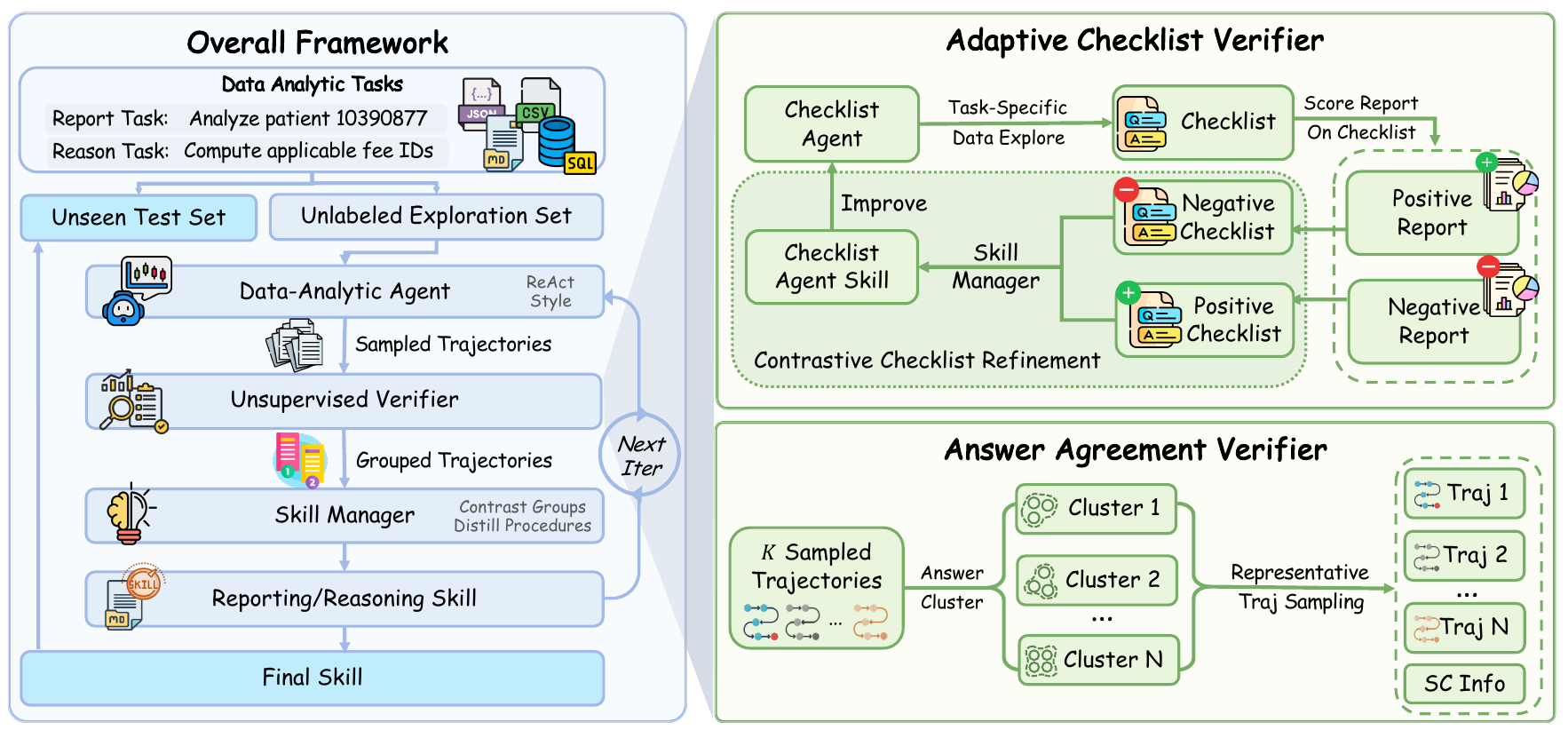}
    \caption{
    \textbf{Overview of the \ours{} framework.}
        The data-analytic agent samples trajectories from an unlabeled exploration set under the current skill, while an unsupervised verifier derives signals and groups trajectories without gold answers or task success labels.
        The Skill Manager contrasts the grouped trajectories to distill reusable procedures and create or update the skill iteratively.
        For report-style tasks, the Adaptive Checklist Verifier uses task-specific checklists to score reports and refine both reporting and checklist-generation skills; for reasoning-style tasks, the Answer Agreement Verifier clusters final answers and estimates self-consistency to guide skill refinement.
    }
    \label{fig:pipeline_v1}
\end{figure*}

Following the \texttt{ReAct} \cite{react} paradigm, the data-analytic agent interleaves reasoning and acting. Before emitting an action at step $t$, the agent generates a thought $z_t$ based on the current context. Consequently, the agent's historical interaction trajectory $h_t$ is formulated as:
\begin{equation}
h_t = (u, z_0, a_0, o_0, z_1, a_1, o_1, \dots, z_{t-1}, a_{t-1}, o_{t-1}).
\end{equation}
Conditioned on the history $h_t$, the agent iteratively predicts the next step via its policy $\pi_\theta$ by $(z_t, a_t) \sim \pi_\theta(\cdot \mid h_t)$ until termination, producing a trajectory $\tau$ and a final answer $y$.

\subsection{LLM Agent Skills}
A skill $\mathcal{S}$ is defined as a structured knowledge bundle that provides reusable procedural guidance for solving specific tasks. Following prior work~\cite{trace2skill}, we represent a skill as:
\begin{equation}
    \mathcal{S} = (\mathcal{M}, \mathcal{R}),
\end{equation}
where $\mathcal{M}$ is a root Markdown document (e.g., \texttt{SKILL.md}), and $\mathcal{R}$ is a set of auxiliary resources. $\mathcal{M}$ typically describes when the skill should be applied, solution strategies, and common failure modes, while $\mathcal{R}$ supports deterministic subtasks.

By injecting a skill, we effectively condition the agent's behavior without altering its underlying parameters $\theta$. At time step $t$, given a task $u$ and the historical context $h_t$, a skill-conditioned agent samples its next action as $(z_t,a_t) \sim \pi_\theta(\cdot \mid h_t, \mathcal{S})$.
The skill guides the agent's interaction with the environment and induces the final answer $y \sim P(y \mid u, \mathcal{S}, \pi_\theta, \mathcal{M}_u)$.
\section{Method}

\subsection{Problem Definition}
\label{sec:task_definition}

We study unsupervised skill discovery for data-analytic agents.
Given an unlabeled exploration set, the goal is to discover a generalizable skill that can improve the agent's performance on unseen data-analysis tasks without using ground-truth answers, success labels, or human annotations.

Let $\mathcal{D}_{\mathrm{explore}}$ denote the unlabeled exploration set used for skill discovery, and let $\mathcal{D}_{\mathrm{test}}$ denote a held-out test set used only for final evaluation.
During the discovery phase, the agent has access only to task inputs, data resources, and interaction trajectories on $\mathcal{D}_{\mathrm{explore}}$.

For a task set $\mathcal{D}$, we define the performance of a skill $\mathcal{S}$ when injected into a data-analytic agent $\pi_\theta$ as
\begin{equation}
J(\mathcal{S}; \pi_\theta, \mathcal{D})
=
\frac{1}{|\mathcal{D}|}
\sum_{u \in \mathcal{D}}
R^\star
\left(
x_T^{u,\mathcal{S}}, u
\right),
\end{equation}
where $x_T^{u,\mathcal{S}}$ denotes the final state or output produced by $\pi_\theta$ on task $u$ under skill $\mathcal{S}$, and $R^\star$ denotes the ground-truth evaluation function used only for offline evaluation.

The unsupervised skill discovery process $\mathcal{E}$ constructs a deployable skill $\widehat{\mathcal{S}}$ from the exploration set while keeping the agent parameters $\theta$ fixed:
\begin{equation}
\widehat{\mathcal{S}}
=
\mathcal{E}
\left(
\mathcal{D}_{\mathrm{explore}}; \pi_\theta
\right).
\end{equation}

The effectiveness of discovery is measured by how well the resulting skill generalizes to unseen tasks.
Ideally, the discovered skill should approach the best skill in the candidate skill space $\mathfrak{S}$ with respect to held-out performance:
\begin{equation}
\widehat{\mathcal{S}}
\approx
\arg\max_{\mathcal{S} \in \mathfrak{S}}
J
\left(
\mathcal{S}; \pi_\theta, \mathcal{D}_{\mathrm{test}}
\right).
\end{equation}
This objective is used only to characterize the desired outcome. $\mathcal{D}_{\mathrm{test}}$ and $R^\star$ are never accessed during skill discovery.

\subsection{Overall Framework}
\label{sec:overall_framework}

Our framework discovers transferable skills through a closed-loop process over unlabeled exploration tasks.
As illustrated in Figure~\ref{fig:pipeline_v1}, the framework consists of three components: a \textit{Data-Analytic Agent} $\pi_\theta$, an \textit{Unsupervised Verifier} $\phi$, and a \textit{Skill Manager} $\psi_\omega$.

At iteration $r$, the Data-Analytic Agent is conditioned on the current skill $\mathcal{S}^{(r)}$, with $\mathcal{S}^{(0)}=\varnothing$.
Following the \texttt{ReAct} paradigm, the agent interacts with task-specific environments and produces exploratory trajectories:
\begin{equation}
\mathcal{T}^{(r)}
=
\left\{
\tau_i^{u,r}
\mid
u \in \mathcal{D}_{\mathrm{explore}},\
i=1,\ldots,N
\right\},
\end{equation}
where $N$ is the number of sampled trajectories per task.

The Unsupervised Verifier analyzes these trajectories without accessing ground-truth answers, success labels, or other privileged supervision.
Instead of directly determining whether a trajectory is correct, the verifier extracts non-privileged signals that characterize trajectory quality, uncertainty, agreement, or divergence:
\begin{equation}
\sigma^{u,r}
=
\phi
\left(
\{\tau_i^{u,r}\}_{i=1}^{N}; u, \mathcal{M}_u
\right),
\end{equation}
where $\mathcal{M}_u$ denotes the task-specific materials available to the agent, such as data files.
These signals provide indirect evidence for distinguishing reliable solution patterns from potentially flawed or incomplete ones.

Based on the verifier signals, trajectories are organized into structured groups:
\begin{equation}
\mathcal{G}^{(r)}
=
\left\{
\mathcal{G}^{(r)}_1,\ldots,\mathcal{G}^{(r)}_K
\right\}.
\end{equation}
Each group corresponds to an unsupervised behavioral pattern, such as relatively high- and low-scoring reports, distinct answer clusters, or trajectories with different self-consistency.

The Skill Manager is implemented as an agentic skill distillation module that can inspect exploration trajectories and the corresponding data files available during exploration. It then contrasts the grouped trajectories and distills reusable procedural knowledge:
\begin{equation}
\mathcal{S}^{(r+1)}
=
\psi_\omega
\left(
\mathcal{S}^{(r)}, 
\mathcal{G}^{(r)}
\right).
\end{equation}
The resulting skill update emphasizes general analysis procedures, robust reasoning strategies, and recurring error-avoidance rules, rather than memorizing task-specific outputs from the exploration set.

The refined skill is injected back into the Data-Analytic Agent, forming an iterative loop of trajectory generation, unsupervised verification, and skill refinement.
Depending on the task type, the verifier signals are instantiated differently.
For report-style tasks, we use checklist-based verification and checklist refinement.
For reasoning-style tasks, we use answer clustering and self-consistency estimation.
After the discovery phase terminates, the final skill $\widehat{\mathcal{S}}$ is fixed and evaluated on $\mathcal{D}_{\mathrm{test}}$ without any further modification.

\subsection{Adaptive Checklist Verifier}
\label{sec:report_verifier}

For report-style data-analytic tasks, the agent is expected to produce a comprehensive analytical report rather than a single fixed answer. 
However, during skill discovery, neither reference reports nor ground-truth checklists are available. 
We therefore introduce an Adaptive Checklist Verifier, which leverages a Checklist Agent to construct task-specific verification criteria without supervision and iteratively refines them together with the report-generation skill.

\paragraph{Task-specific Checklist Generation}
Given a task $u$, the Checklist Agent generates a task-specific checklist 
$\mathcal{C}^{u}=\{c_1,\ldots,c_L\}$. 
Each checklist item is formulated as a checkable question--answer criterion tailored to the task, specifying an analytical insight or requirement that the report is expected to address. 
For a report $y_i^{u,r}$ generated by the Data-Analytic Agent at round $r$, the verifier assigns a checklist score:
\begin{equation}
q_i^{u,r}
=
\mathrm{Score}(y_i^{u,r}, \mathcal{C}^{u})
=
\frac{1}{|\mathcal{C}^{u}|}
\sum_{c \in \mathcal{C}^{u}}
s(y_i^{u,r}, c),
\end{equation}
where $s(y_i^{u,r}, c)\in[0,1]$ measures the extent to which the report satisfies checklist item $c$. 
The resulting score provides an unsupervised quality signal, rather than a ground-truth correctness label.

\paragraph{Report-Side Skill Evolution}
Based on the checklist scores, the verifier partitions the sampled report generation trajectories and according checklists into a relatively positive group $\mathcal{G}^{(r)}_{+}$ and a relatively negative group $\mathcal{G}^{(r)}_{-}$ according to the average score over all tasks in the current round. 
The Skill Manager then contrasts these two groups to update the report-generation skill:
\begin{equation}
\mathcal{S}^{(r+1)}_{\pi}
=
\psi_\omega
\left(
\mathcal{S}^{(r)}_{\pi},
\mathcal{G}^{(r)}_{+},
\mathcal{G}^{(r)}_{-}
\right).
\end{equation}
This update distills reusable strategies from high-scoring reports and suppresses recurring weaknesses observed in low-scoring ones.
The report-side skill evolution continues until the average report score on the generated checklists decreases.

\paragraph{Contrastive Checklist Refinement}
A key challenge is that a static checklist $\mathcal{C}^{u}$ may fail to capture all relevant information required for high-quality reporting. Therefore, the Data-Analytic Agent may overfit to the checklist and improve its verifier score without genuinely improving report quality.
To mitigate this verifier overfitting problem, we introduce a contrastive checklist refinement stage.

Once the average score on $\mathcal{D}_{\mathrm{explore}}$ decreases, we redirect the Skill Manager's optimization target from the Data-Analytic Agent's skill $\mathcal{S}_{\pi}$ to the Checklist Agent's skill $\mathcal{S}_{\phi}$. During this stage, we utilize checklist generation trajectories and according reports to refine the checklist agent.
Specifically, high-scoring reports are leveraged as contrastive cases to identify checklist omissions, whereas low-scoring reports provide evidence of checklist dimensions that effectively detect report weaknesses.

Accordingly, the Skill Manager updates the checklist-side skill $\mathcal{S}_{\phi}$ by reversing the contrastive direction:
\begin{equation}
\mathcal{S}^{(r'+1)}_{\phi}
=
\psi_\omega
\left(
\mathcal{S}^{(r')}_{\phi},
\mathcal{G}^{(r')}_{-},
\mathcal{G}^{(r')}_{+}
\right),
\end{equation}
where $r'$ indexes the refinement loop of the checklist agent. 
With the reports fixed, the checklist-generation skill is iteratively refined until the average checklist score over these reports ceases to decrease. The checklist generated by the resulting skill is subsequently used to support the next round of report-side skill evolution.
Through this alternating process, the Checklist Agent becomes progressively more discriminative, while the Data-Analytic Agent is driven to produce more comprehensive analytical reports.

\definecolor{sevRowBg}{HTML}{EEF2FA}
\definecolor{skillcreatorbg}{HTML}{EEF9F8} 
\definecolor{oursbg}{HTML}{F4F8FF}         

\providecommand{\unc}[2]{#1$^{\text{\scriptsize\color{blue!45!cyan}$\uparrow$\normalcolor\ensuremath{#2}}}$}

\begin{table*}[ht]
    \centering
    \caption{Performance comparison on Deep Data Research.
    The highest score in each column is shown in \textbf{bold}, while the second-highest score is \underline{underlined}.
    Accuracy measures the fraction of checklist items that can be verified from the insights extracted by the model, with results reported under both sample-level averaging over task entities and item-level averaging over checklist items.}
    \label{tab:all_databases}
    \newcolumntype{C}{>{\centering\arraybackslash}p{1.0cm}}
    \resizebox{0.98\textwidth}{!}{%
    \begin{tabular}{l|*{12}{C}c}
    \toprule
    & \multicolumn{6}{c|}{\textit{Sample-Averaged Accuracy}} & \multicolumn{6}{c}{\textit{Item-Averaged Accuracy}} & \multicolumn{1}{c}{} \\
    \cmidrule(lr){2-7} \cmidrule(lr){8-13}
    \multirow{2}{*}{\textbf{Models}} 
    & \multicolumn{3}{c|}{\textit{Message}} & \multicolumn{3}{c|}{\textit{Trajectory}} 
    & \multicolumn{3}{c|}{\textit{Message}} & \multicolumn{3}{c|}{\textit{Trajectory}} 
    & \multirow{2}{*}{\textit{Overall Avg.}} \\
    \cmidrule(lr){2-4} \cmidrule(lr){5-7} \cmidrule(lr){8-10} \cmidrule(lr){11-13}
    & MIMIC & GLOBEM & 10-K & MIMIC & GLOBEM & 10-K & MIMIC & GLOBEM & 10-K & MIMIC & GLOBEM & 10-K & \\
    \midrule
    \multicolumn{14}{c}{\textit{\textbf{No Skill}}} \\
    \midrule
    \claudemoji{}~Claude 4.6 Sonnet & 36.18 & 60.64 & \best{77.00} & 37.05 & 58.92 & \second{66.34} & 34.26 & 60.62 & \best{76.92} & 35.12 & 59.08 & \second{66.41} & 55.71 \\        
    \claudemoji{}~Claude 4.5 Sonnet & 34.90 & 56.54 & 75.32 & 34.92 & 56.18 & 54.01 & 32.53 & 56.62 & 75.35 & 33.04 & 56.31 & 54.47 & 51.68 \\
    \Openaiemoji{}~GPT-5-2 & 37.20 & 49.93 & 60.16 & 37.07 & 59.63 & 48.52 & 35.12 & 49.85 & 59.65 & 34.78 & 59.69 & 48.51 & 48.34  \\
    \deepseekemoji{}~DeepSeek-V4-Pro & 30.51 & 55.76 & 67.51 & 35.61 & 60.88 & 57.72 & 28.72 & 56.00 & 67.66 & 33.56  & 60.92 & 58.08 & 51.08 \\       
    \Qwenemoji{}~Qwen3.5-397B-A17B & 30.49 & 49.78 & 50.02 & 30.04 & 45.07 & 27.99 & 28.20 & 49.54 & 49.29 & 28.20 & 44.62 & 28.10 & 38.45 \\
    \midrule
    \multicolumn{14}{c}{\textit{\textbf{Skill Creator}}} \\
    \midrule
    \rowcolor{skillcreatorbg}
    \claudemoji{}~Claude 4.5 Sonnet & 36.09 & 58.46 & 70.50 & 40.54 & 62.13 & 57.58 & 34.08 & 58.15 & 70.96 & 38.58 & 61.85 & 57.93 & \unc{53.90}{2.22}  \\
    \rowcolor{skillcreatorbg}
    \Openaiemoji{}~GPT-5-2 & \second{37.59} & 53.85 & 63.15 & 35.28 & 70.64 & 48.80 & \second{35.47} & 53.54 & 63.73 & 33.56 & 70.46 & 48.35 & \unc{51.20}{2.86}  \\
    \rowcolor{skillcreatorbg}
    \deepseekemoji{}~DeepSeek-V4-Pro & 28.32 & 59.00 & 69.84 & \second{43.27} & 65.81 & 58.46 & 25.78 & 59.38 & 69.86 & \second{40.66} & 65.85 & 58.71 & \unc{53.75}{2.67} \\
    \rowcolor{skillcreatorbg}   
    \Qwenemoji{}~Qwen3.5-397B-A17B & 32.63 & 53.33 & 58.57 & 37.45 & 58.50 & 40.68 & 30.97 & 53.23 & 58.56 & 35.12 & 58.15 & 40.97 & \unc{46.51}{8.06} \\
    \midrule
    \multicolumn{14}{c}{\textit{\textbf{\ours}}} \\
    \midrule
    \rowcolor{oursbg}
    \claudemoji{}~Claude 4.5 Sonnet & 36.75 & 58.68  & 76.55 & 36.99 & \second{72.92} & 65.63 & 34.60 & 58.77 & \second{76.45} & 34.08 & \second{72.92} & 65.62 & \unc{57.50}{5.82}  \\
    \rowcolor{oursbg}   
    \Openaiemoji{}~GPT-5-2 & \best{39.15} & \second{62.21} & 65.97 & 39.77 & 70.56 & 55.94 & \best{37.20} & \second{62.15} & 65.93 & 37.72 & 70.46 & 55.57 & \unc{55.22}{6.88}  \\   
    \rowcolor{oursbg}
    \deepseekemoji{}~DeepSeek-V4-Pro & 32.18 & 57.79 & \second{76.69} & \best{43.77}  & 66.42 & \best{71.54} & 30.62 & 57.85 & 76.30 & \best{42.39} & 66.46 & \best{71.90} & \unc{\second{57.83}}{6.75} \\
    \rowcolor{oursbg}
    \Qwenemoji{}~Qwen3.5-397B-A17B & 35.78 & \best{64.02} & 68.22 & 41.50 & \best{73.95} & 65.56 & 33.56 & \best{63.69} & 68.45 & 39.27 & \best{73.85} & 66.25 & \unc{\best{57.84}}{19.39}\\
    \bottomrule
    \end{tabular}%
    }
\end{table*}

\definecolor{mygrey}{RGB}{213, 213, 213}
\newcommand{\GG}{\cellcolor{mygrey}}
\definecolor{sevRowBg}{HTML}{EEF2FA}
\definecolor{skillcreatorbg}{HTML}{EEF9F8} 
\definecolor{oursbg}{HTML}{F4F8FF}         

\providecommand{\unc}[2]{#1$^{\text{\scriptsize\color{blue!45!cyan}$\uparrow$\normalcolor\ensuremath{#2}}}$}

\providecommand{\dec}[2]{#1$^{\text{\scriptsize\color{red!60!orange}$\downarrow$\normalcolor\ensuremath{#2}}}$}

\providecommand{\same}[2]{#1$^{\text{\scriptsize\color{gray}$\rightarrow$\normalcolor\ensuremath{#2}}}$}

\begin{table}
\footnotesize
\renewcommand\arraystretch{1.1}
\caption{Performance comparison on DABStep. The highest score in each column is shown in \textbf{bold}, while the second-highest score is \underline{underlined}.}
\centering
\scalebox{1.0}{
\begin{tabular}{l|cc|c}
\toprule
\textbf{Models} &
\textbf{Easy} &
\textbf{Hard} & 
\textbf{All} \\
\midrule
\multicolumn{4}{c}{\textit{No Skill}} \\
\midrule
Claude Sonnet 4.6 & 80.86 & 33.57 & 41.13 \\
Claude Sonnet 4.5 & \second{88.89} & 27.35 & 37.18 \\
GPT-5-2 & 85.19 & 16.78 & 27.71 \\
DeepSeek-V4-Pro & 80.25 & 11.97 & 13.70 \\
Qwen3.5-397B & 80.86 & 29.81 & 37.97 \\
\midrule
\multicolumn{4}{c}{\textit{Skill Creator}} \\
\midrule
\rowcolor{skillcreatorbg}
Claude Sonnet 4.5 & \dec{81.48}{7.41} & \unc{47.89}{20.54} & \unc{53.26}{16.08} \\
\rowcolor{skillcreatorbg}
GPT-5-2 & \unc{85.19}{0.00} & \unc{46.36}{29.58} & \unc{52.56}{24.85} \\
\rowcolor{skillcreatorbg}
DeepSeek-V4-Pro & \unc{82.10}{1.85} & \unc{39.91}{27.94} & \unc{46.65}{32.95} \\
\rowcolor{skillcreatorbg}
Qwen3.5-397B & \dec{78.40}{2.46} & \unc{49.88}{20.07} & \unc{54.44}{16.47} \\
\midrule
\multicolumn{4}{c}{\textit{\ours}} \\
\midrule
\rowcolor{oursbg}
Claude Sonnet 4.5 & \dec{87.04}{1.85} & \unc{\second{57.40}}{30.05} & \unc{\second{62.13}}{24.95} \\
\rowcolor{oursbg}
GPT-5-2 & \unc{\best{90.12}}{4.93} & \unc{56.45}{39.67} & \unc{61.83}{34.12} \\
\rowcolor{oursbg}
DeepSeek-V4-Pro & \unc{87.66}{7.41} & \unc{53.52}{41.55} & \unc{58.97}{45.27} \\
\rowcolor{oursbg}
Qwen3.5-397B & \unc{87.04}{6.18} & \unc{\best{58.22}}{28.41} & \unc{\best{62.82}}{24.85} \\
\bottomrule
\end{tabular}
}
\label{tab:agent_results}
\end{table}

\subsection{Answer Agreement Verifier}
\label{sec:reasoning_verifier}

For reasoning-style tasks requiring fixed answers, gold labels are unavailable during exploration. To address this, we propose a label-free Answer Agreement Verifier that evaluates trajectories via answer-level clustering and self-consistency. Rather than predicting correctness, it groups trajectories to uncover stable solution patterns and divergent reasoning behaviors.

\paragraph{Answer Clustering}
For task $u$ at iteration $r$, the agent generates $N$ trajectories with final answers $\{y_i^{u,r}\}_{i=1}^{N}$. We apply a clustering operator that partitions these answers based on a type-specific equality metric (e.g., exact match).

\paragraph{Self-Consistency Estimation}
The self-consistency (SC) score of a trajectory is then defined by the relative size of its assigned answer cluster.
While SC measures answer convergence, it does not guarantee correctness.
Therefore, the verifier leverages SC merely as an auxiliary uncertainty signal.

\paragraph{Agent-Side Skill Evolution}
The verifier organizes trajectories into structured groups $\mathcal{G}^{(r)}$ based solely on their answer clusters, with SC appended as an auxiliary confidence signal.
Each answer cluster contains trajectories that converge to the same final answer, and we represent each cluster with a single representative trajectory selected by prioritizing fewer interaction turns and fewer execution exceptions.
In this way, the grouped trajectories provide both cluster-level agreement signals and concise representative reasoning traces.
During iterative refinement, we further filter out trajectories whose SC remains saturated across consecutive iterations, so that the Skill Manager can focus on less certain cases that still require refinement.
The Skill Manager then compares representatives across the remaining clusters to identify divergent reasoning behaviors and recurring failure modes, and uses these contrastive signals to update the agent-side skill.
By distilling reusable reasoning strategies from representative trajectories and cross-cluster differences, the agent iteratively improves its reasoning robustness in an unsupervised manner.
\begin{figure*}[t] 
    \centering
    \begin{subfigure}[b]{0.32\textwidth} 
        \centering
        \includegraphics[width=\textwidth]{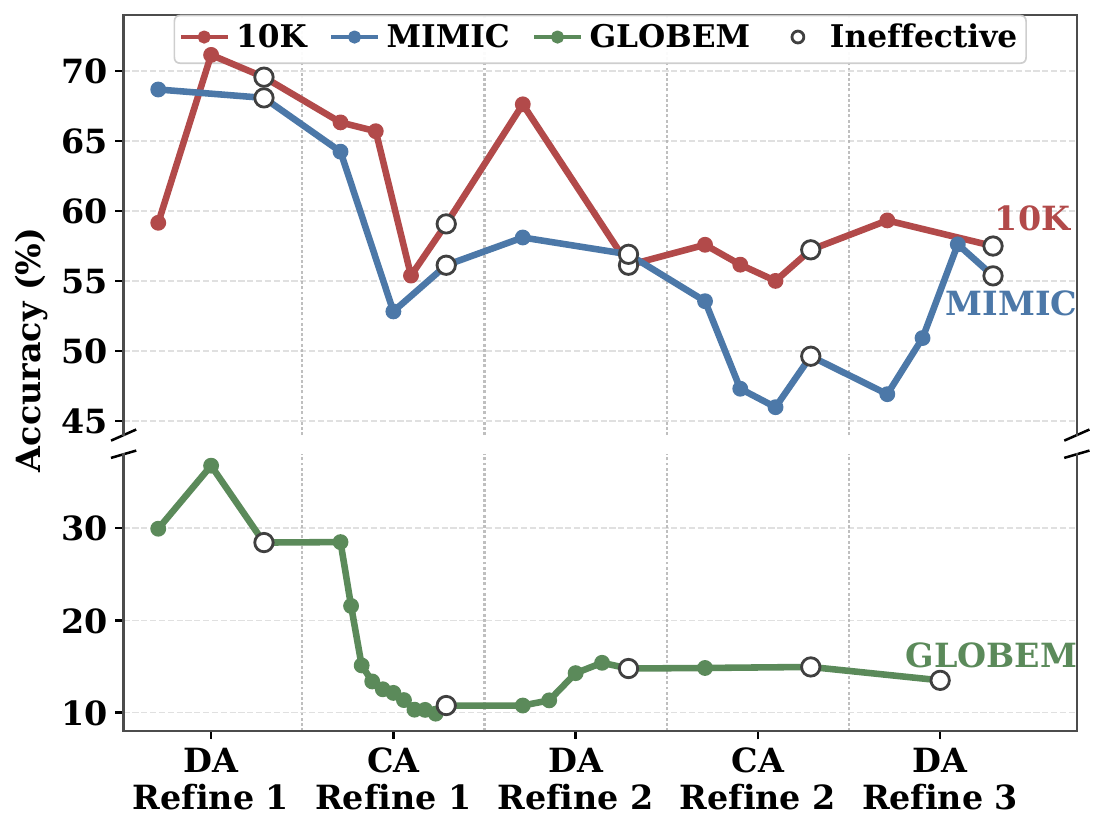}
        \caption{Checklist-Score Dynamics on Refinement.} 
        \label{fig:report_checklist_agent_iter}
    \end{subfigure}
    \begin{subfigure}[b]{0.32\textwidth}
        \centering
        \includegraphics[width=1.0\textwidth]{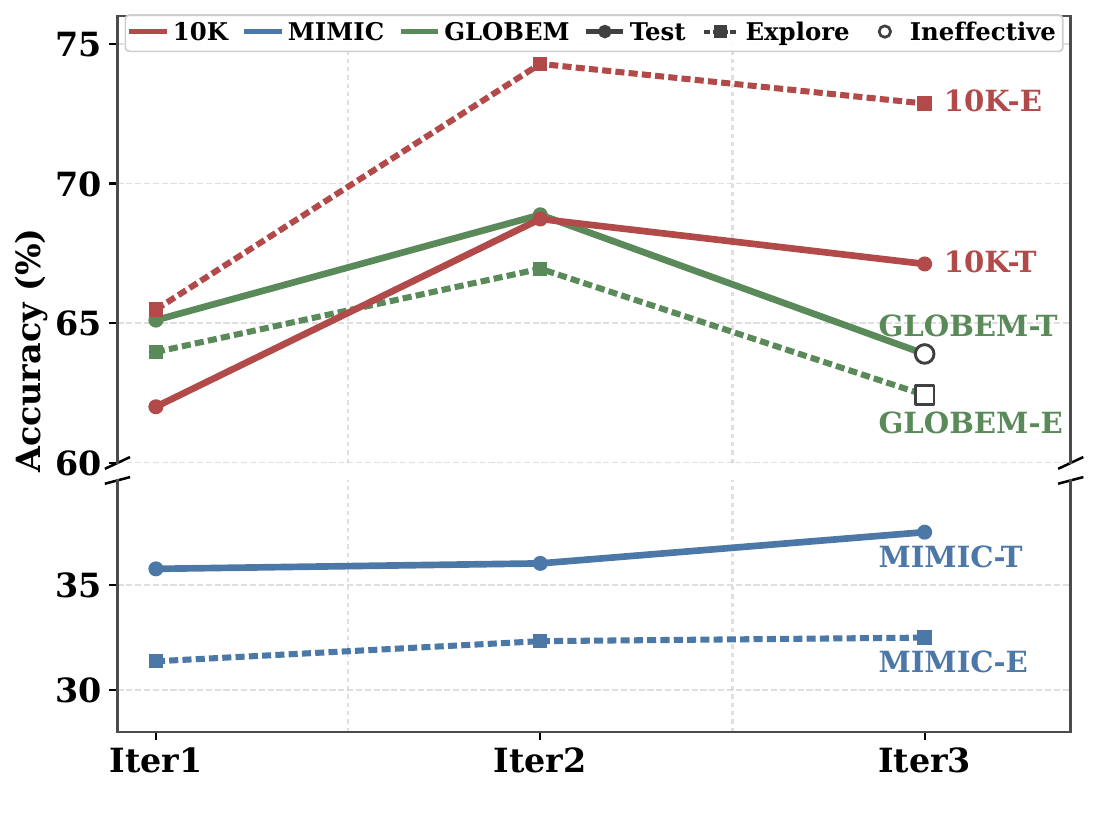}
        \caption{Report-Task Iteration Analysis.}
        \label{fig:report_iter}
    \end{subfigure}
    \begin{subfigure}[b]{0.34\textwidth}
        \centering
        \includegraphics[width=\textwidth]{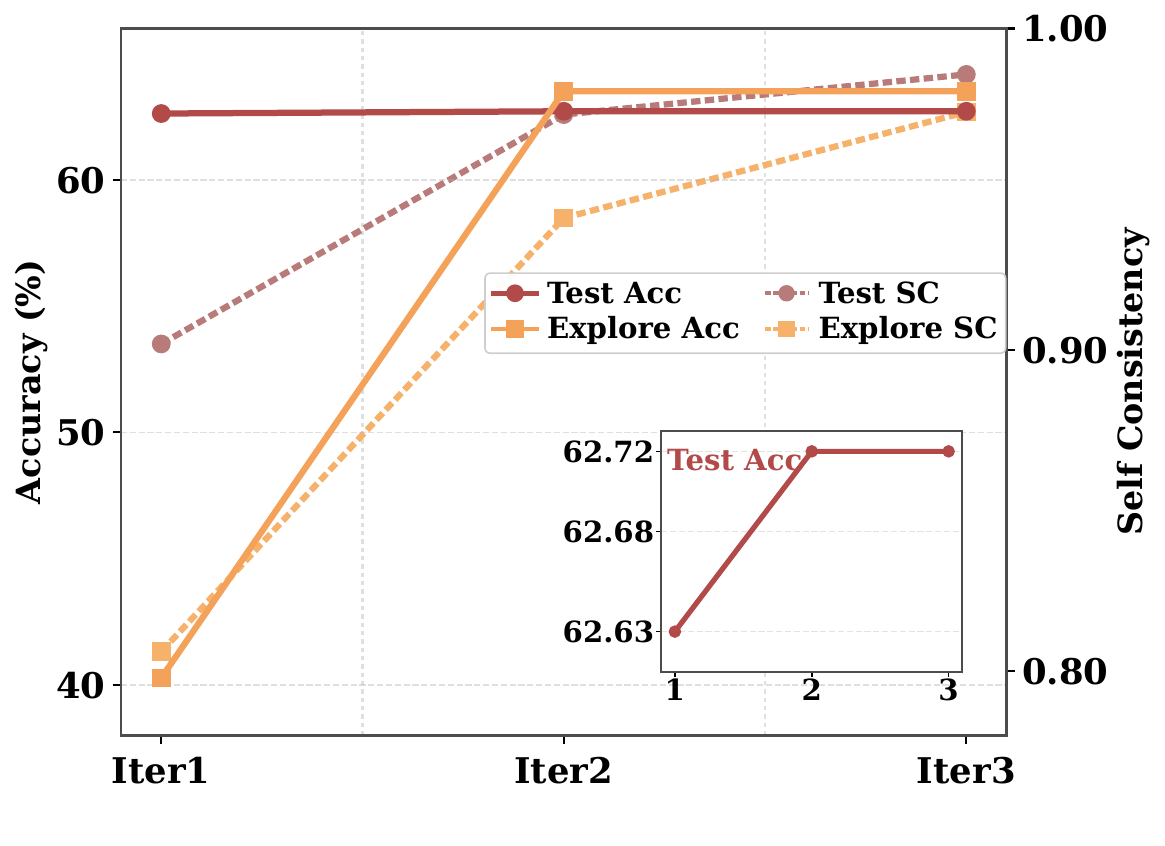}
        \caption{Reasoning-Task Iteration Analysis.}
        \label{fig:reason_iter}
    \end{subfigure}

    \caption{\textbf{Iteration Analysis.}
    (a) \textbf{Checklist-Score Dynamics on Refinement.}
    We track the scores of reports generated by the Data-Analytic Agent on checklists generated by the Checklist Agent throughout the refinement process. Hollow markers denote refinement steps that fail to produce a valid skill update.
    (b) \textbf{Report-Task Iteration Analysis.}
    We evaluate the last valid skill after each Data-Analytic Agent refinement round using ground-truth checklists on the explore and test sets.
    (c) \textbf{Reasoning-Task Iteration Analysis.}
    We report the accuracy and self-consistency of the reasoning-task agent on the explore and test sets across refinement iterations.}
\end{figure*}

\section{Experiment}

\subsection{Experimental Settings}
\label{sec:experiment_settings}

\paragraph{Benchmarks and Metrics}
We conduct our evaluation on two distinct categories of data analysis benchmarks: the report-based data analysis task, Deep Data Research~\cite{ddr}, and the reasoning-based data analysis task, DABStep~\cite{dabstep}.
For both benchmarks, we randomly partition the instances into $\mathcal{D}_\mathrm{explore}$ and $\mathcal{D}_\mathrm{test}$ at a 1:3 ratio.
For Deep Data Research, we report the sample-averaged accuracy and item-averaged accuracy in original paper, and specifically detail the message-wise insights and trajectory-wise insights. We use GPT-5-mini~\cite{gpt-5} as the judge model.
For DABStep, we report the overall accuracy.

\paragraph{Baselines and Models}
We compare our approach against Anthropic's Skill Creator~\cite{anthropic2026skillcreator}.
We implement the baseline with Claude Code equipped with Skill Creator skill, which creates skills by exploring the Data-Analytic Agent's trajectories on $\mathcal{D}_\mathrm{explore}$ under the same data access privileges as our method.
To assess the effectiveness of our framework, we evaluate a diverse suite of base models that vary significantly in parameter scale and reasoning paradigms. 
Specifically, our evaluation includes Claude-Sonnet-4.6~\cite{anthropic2026claudesonnet46systemcard}, Claude-Sonnet-4.5~\cite{anthropic2025claudesonnet45systemcard}, GPT-5.2 (medium reasoning mode)~\cite{gpt-5}, DeepSeek-V4-Pro (non-reasoning mode)~\cite{deepseek-v4}, as well as Qwen3.5-397B-A17B (non-reasoning mode)~\cite{qwen35}.

\paragraph{Implementation Details}
Regarding specific implementation details, the baseline Skill-Creator is implemented using Claude Code powered by Claude Sonnet 4.6. For our proposed method, both the Data-Analytic Agents and the Checklist Agent utilize Qwen3.5-397B-A17B, while the Skill Manager also employs Claude Code powered by Claude Sonnet 4.6. During the exploration phase, the sampling strategy for the Data-Analytic Agents varies by task type. We sample exactly one trajectory per instance for the Reporting tasks, and ten trajectories per instance for the Reasoning tasks.
For Reporting tasks, our skill iteration alternates between updating the Data-Analytic Agent and updating the Checklist Agent, resulting in three Data-Analytic Agent updates interleaved with two Checklist Agent updates.
For Reasoning tasks, our skill iteration directly keeps three.
For skill granularity, each DDR subset uses a separate skill, and DABStep tasks are divided into nine categories with one skill per category. The same setting is used for the Skill-Creator baseline.
Finally, the sampling temperature for the Data-Analytic Agents is set to 1.0 during exploration and is adjusted to 0.0 during evaluation.

\definecolor{skillbg}{HTML}{F4F8FF}

\begin{table}[t]
\centering
\caption{
\textbf{Ablation study of the reporting verifier on 10-K.}
TS denotes task-specific checklist generation, 
and CR denotes iterative checklist refinement.
The best result is shown in \textbf{bold}, and the second-best result is \underline{underlined}.
}
\label{tab:reporting_ablation}
\renewcommand\arraystretch{1.0}
\begin{tabular}{l|cc|cc|c}
\toprule
\textbf{Variant} 
& \textbf{TS}
& \textbf{CR}
& \textbf{Sample}
& \textbf{Item}
& \textbf{10-K} \\
\midrule
\rowcolor{skillbg}
\textbf{Ours} 
& \cmark & \cmark & \best{66.89} & \best{67.35} & \best{67.12} \\
w/o Checklist Refinement
& \cmark & \xmark & \second{61.46} & \second{62.56} & \second{62.01} \\
w/o Task-Specific Checklist
& \xmark & \cmark & 52.30 & 52.12 & 52.21 \\
w/o Checklist Agent
& \xmark & \xmark & 53.51 & 53.14 & 53.32 \\
\bottomrule
\end{tabular}
\end{table}
\definecolor{skillbg}{HTML}{F4F8FF}

\begin{table}[t]
\centering
\caption{
\textbf{Ablation study of the reasoning verifier on DABStep.}
AC denotes answer clustering, and SC denotes self-consistency.
The best result is shown in \textbf{bold}, and the second-best result is \underline{underlined}.
}
\label{tab:reasoning_ablation}
\renewcommand\arraystretch{1.1}
\begin{tabular}{l|cc|cc|c}
\toprule
\textbf{Variant} 
& \textbf{AC} 
& \textbf{SC}
& \textbf{Easy}
& \textbf{Hard}
& \textbf{All} \\
\midrule
\rowcolor{skillbg}
\textbf{Ours} 
& \cmark & \cmark & \second{87.04} & \best{58.22} & \best{62.82}
 \\
w/o Self-Consistency 
& \cmark & \xmark & \best{88.89} & \second{49.65} & \second{55.92} \\
w/o Answer Clustering 
& \xmark & \cmark & 85.19 & 40.85 & 47.93 \\
All Trajectories 
& \xmark & \xmark & 85.19 & 47.89 & 53.85 \\
\bottomrule
\end{tabular}
\end{table}

\begin{figure*}[t] 
    \centering
    \begin{subfigure}[b]{0.33\textwidth} 
        \centering
        \includegraphics[width=\textwidth]{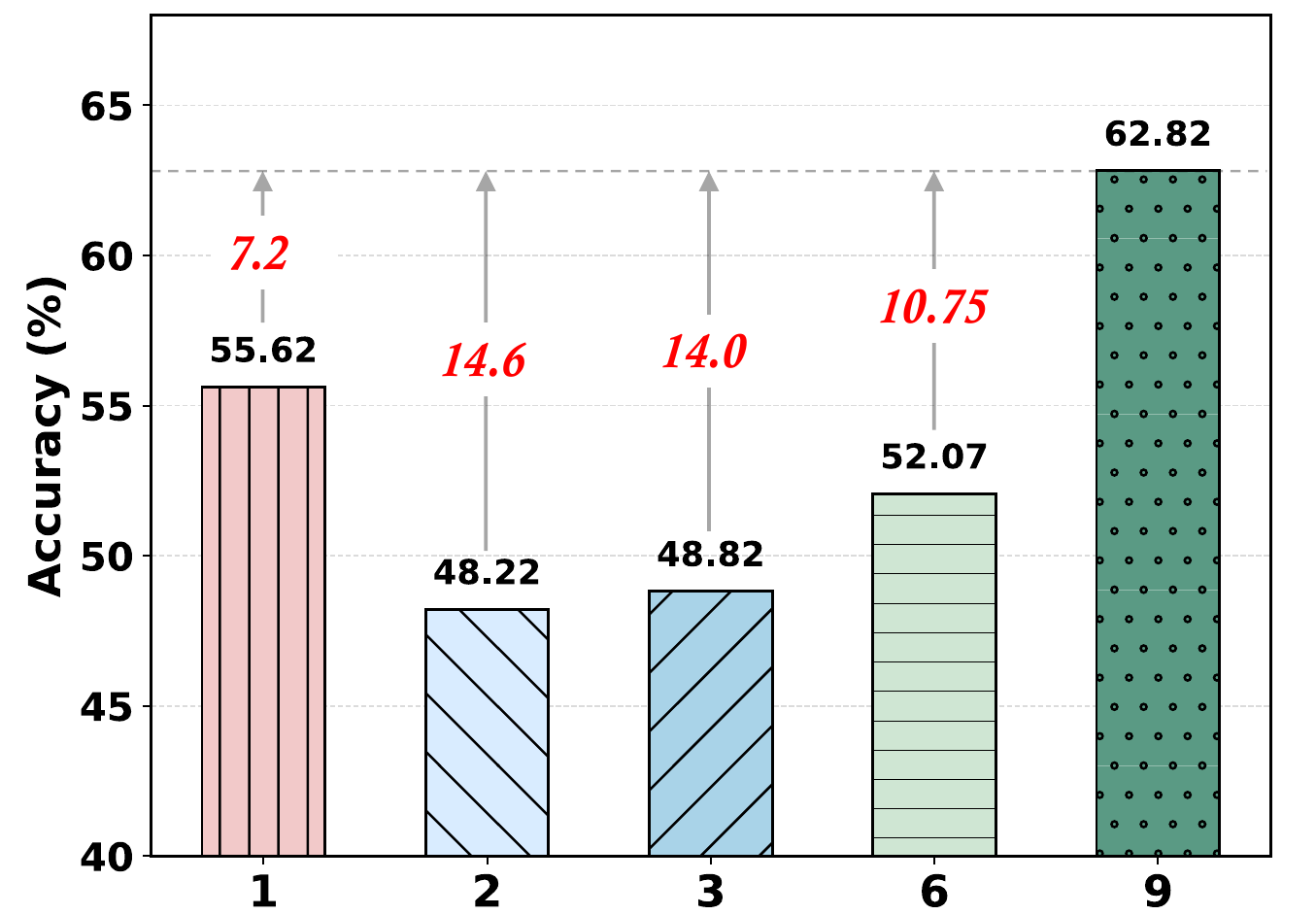}
        \caption{Skill Granularity Analysis.} 
        \label{fig:skill_granularity}
    \end{subfigure}
    \begin{subfigure}[b]{0.32\textwidth} 
        \centering
        \includegraphics[width=1.0\textwidth]{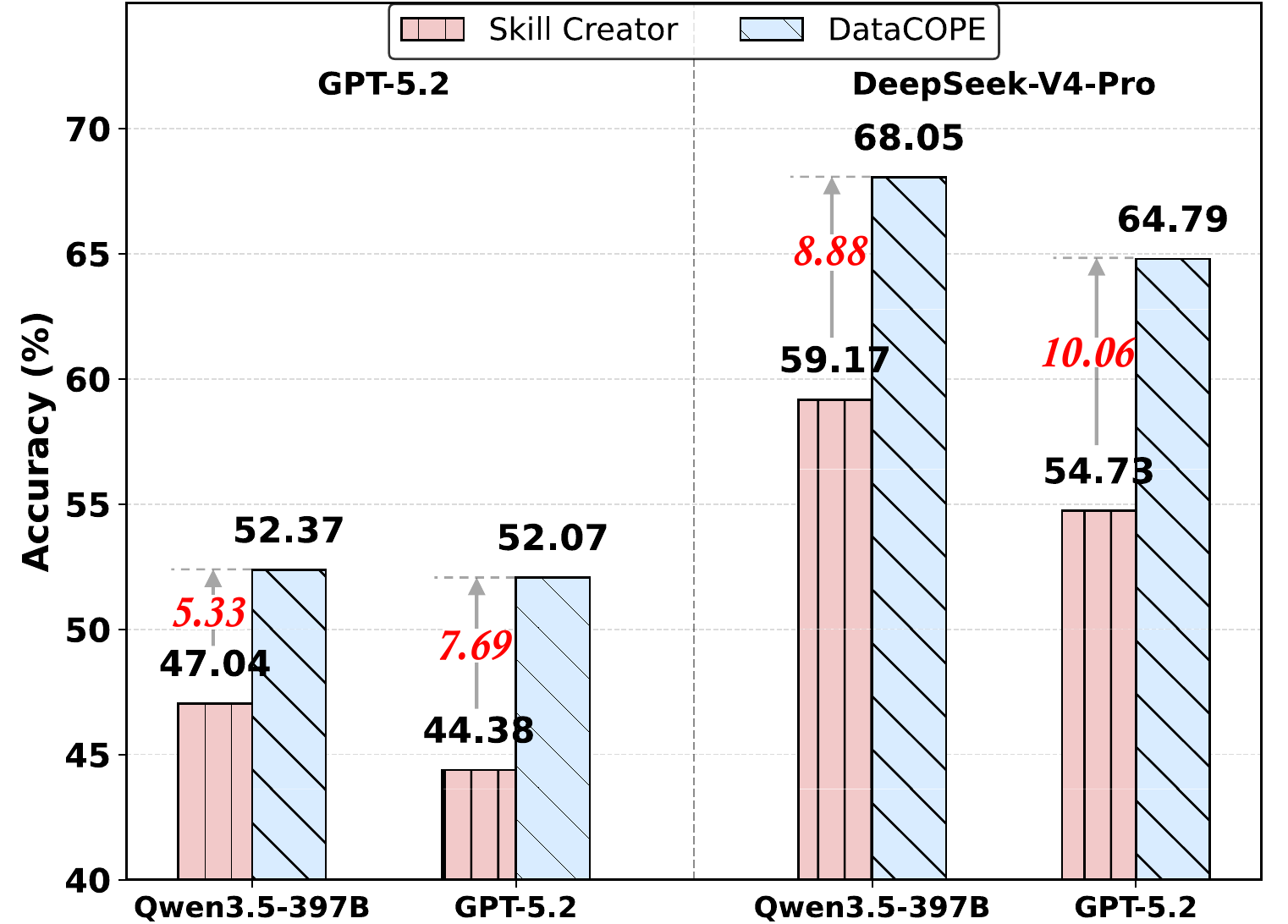}
        \caption{Data-Analytic Agent Analysis.} 
        \label{fig:data_agent_analysis}
    \end{subfigure}
    \begin{subfigure}[b]{0.32\textwidth}
        \centering
        \includegraphics[width=\textwidth]{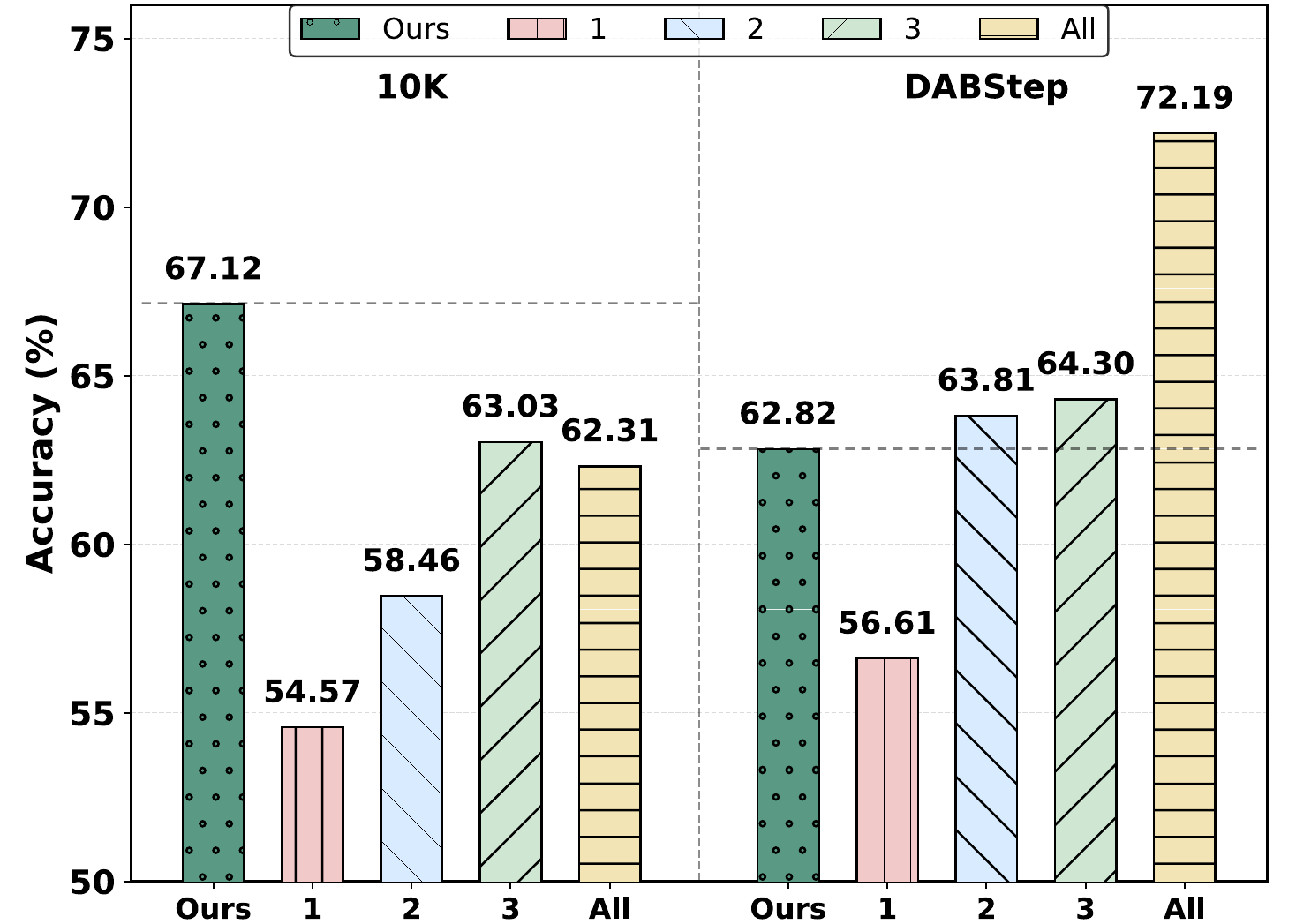}
        \caption{Supervised Skill Discovery Analysis.}
        \label{fig:supervised_analysis}
    \end{subfigure}

    \caption{\textbf{Further Analysis of \ours{}.}
    (a): \textbf{Skill Granularity Analysis}. We evaluate different skill granularities on DABStep and show that proper granularity is crucial for effective skill discovery.
    (b): \textbf{Data-Analytic Agent Analysis}. We replace the Data-Analytic Agent in \ours{} and find that \ours{} consistently improve the performance of skill discovery.
    (c): \textbf{Supervised Skill Discovery Analysis}. We compare \ours{} with Skill Creator using randomly labeled trajectories, demonstrating that \ours{} achieves competitive performance with zero annotation cost.}
\end{figure*}

\subsection{Main Results}
\paragraph{\textbf{\ours{} Consistently Improves Agents across Task Formats}}
Tables~\ref{tab:all_databases} and~\ref{tab:agent_results} show that \ours{} consistently improves data-analytic agents on both reporting and reasoning tasks.
On reporting tasks, \ours{} raises the mean Overall Avg. from 47.39\% to 57.10\% across the four matched base models.
On reasoning tasks, \ours{} brings larger gains on DABStep, especially on the hard split, increasing the mean score from 29.14\% to 61.44\%.
These results demonstrate that the discovered skills are not limited to a single task format, but can benefit both open-ended report generation and fixed-answer data reasoning.

\paragraph{\textbf{\ours{} Transfers across Different Base Models}}
The improvement of \ours{} is consistent across different model families.
On reporting tasks, all four matched models benefit from the discovered skills, with Qwen3.5-397B obtaining the largest gain and achieving the best performance.
On reasoning tasks, the same trend holds across Claude, GPT, DeepSeek, and Qwen models, with particularly strong improvements on hard DABStep instances.
This consistent cross-model improvement indicates that \ours{} captures generalizable data-analysis procedures rather than overfitting to model-specific prompting patterns.

\paragraph{\textbf{Unsupervised Verifier Signals Enable More Effective Skill Discovery}}
Although Skill Creator improves base agents in many cases, its gains are consistently smaller than those of \ours{}.
On reporting tasks, Skill Creator raises the mean Overall Avg. to 51.34\%, whereas \ours{} further improves it to 57.10\%.
On reasoning tasks, Skill Creator reaches a mean  all score of 51.73\%, while \ours{} achieves 61.44\%.
These results indicate that the advantage of \ours{} stems not merely from skill generation itself, but from the unsupervised verifier signals, which enables more effective discovery of transferable data-analysis skills.

\subsection{Analysis}
\paragraph{\textbf{Iterative Skill Refinement Is Effective but Not Monotonic}}
For reporting tasks, we first examine the checklist-score dynamics during refinement in Figure~\ref{fig:report_checklist_agent_iter}.
The scores suggest that the Data-Analytic Agent obtains more effective improvements in the early refinement stage, while later refinements become less effective or even invalid for 10-K and GLOBEM.
In contrast, MIMIC continues to benefit from later refinement.
We further analyze the explore and test performance of the refined skills in Figure~\ref{fig:report_iter}.
The second iteration consistently improves performance across all datasets, whereas the third iteration is not uniformly beneficial. 10-K and GLOBEM show diminishing or negative gains in later iterations, while MIMIC still improves in the final report evaluation.
This trend is consistent with the checklist-score dynamics, suggesting that checklist-based scores can serve as useful verifier-side diagnostic signals for identifying effective refinements and filtering invalid skill updates.

For reasoning tasks, as shown in Figure~\ref{fig:reason_iter}, refinement mainly improves consistency.
Self-consistency increases substantially on both explore and test splits, but test accuracy remains nearly unchanged.
This suggests that answer-level verification reduces variance but may fail when the dominant answer cluster is incorrect.

\paragraph{\textbf{Verifier Components are Critical for Skill Discovery}}
For the reporting task on 10-K, Table~\ref{tab:reporting_ablation} shows that removing the Checklist Agent decreases the score from 67.12\% to 53.32\%, indicating that trajectory-level exploration alone provides insufficient feedback for effective skill discovery.
Replacing task-specific checklists with generic ones further reduces the score to 52.21\%, suggesting that non-adaptive verification criteria can introduce noisy supervision.
In addition, removing checklist refinement lowers the score to 57.30\%, demonstrating that iterative refinement of the Checklist Agent is important for producing informative feedback.

For the reasoning task on DABStep, Table~\ref{tab:reasoning_ablation} shows that using all trajectories without verifier-based selection reduces the score from 62.82\% to 53.85\%, suggesting that raw trajectories alone are insufficient.
Removing answer clustering causes the largest degradation, with the score dropping to 47.93\%, even below the all-trajectory variant.
This indicates that relying solely on self-consistency can be harmful, since multiple trajectories may converge to the same incorrect answer.
Removing self-consistency also decreases the score to 55.92\%, showing that self-consistency remains beneficial when combined with answer clustering.
Overall, answer clustering provides the primary signal for mitigating misleading consensus, while self-consistency further estimates the reliability of each answer cluster.

\definecolor{skillbg}{HTML}{F4F8FF}
\newcommand{\gain}[1]{\textcolor{blue!60!black}{\scriptsize $\uparrow$#1}}
\newcommand{\save}[1]{\textcolor{green!45!black}{\scriptsize $\downarrow$#1}}

\newcommand{\skillcell}[1]{\cellcolor{skillbg}{#1}}

\begin{table}[t]
\centering
\small
\setlength{\tabcolsep}{4.5pt}
\renewcommand{\arraystretch}{1.12}
\caption{
\textbf{Efficiency and effectiveness of the discovered skill.}
We compare the average token usage per task and task accuracy of different agents with and without the discovered skill.
The skill substantially reduces token consumption while improving accuracy.
}
\label{tab:skill_efficiency}
\resizebox{\linewidth}{!}{
\begin{tabular}{lcrrrr}
\toprule
\textbf{Model} & \textbf{Agent Scaffold} & \textbf{Setting} 
& \textbf{Avg. Tokens} 
& \textbf{Acc.} 
& \textbf{Token Saving} \\
\midrule
\multirow{2}{*}{Sonnet 4.6}
& \multirow{2}{*}{Claude Code}
& w/o Skill 
& 241{,}275 
& 44.00 
& --  \\
& & \skillcell{w/ Skill}
& \skillcell{\textbf{64{,}157}}
& \skillcell{\textbf{64.00}}
& \skillcell{\save{73.4\%}} \\
\midrule
\multirow{2}{*}{Qwen3.5-397B}
& \multirow{2}{*}{ReAct}
& w/o Skill 
& 110{,}116 
& 36.00 
& -- \\
& & \skillcell{w/ Skill}
& \skillcell{\textbf{64{,}213}} 
& \skillcell{\textbf{62.00}} 
& \skillcell{\save{41.7\%}}  \\
\bottomrule
\end{tabular}
}
\end{table}
\section{Further Analysis}
\subsection{Skill Granularity Analysis}
We study how the granularity of the discovered skills affects the performance on DABStep.
As shown in Fig.~\ref{fig:skill_granularity}, using all 9 discovered skills achieves the best performance, reaching 62.82\% accuracy.
In contrast, using 2 or 3 skills performs even worse than using a single skill. The skills are no longer general enough to provide broad guidance, yet remain too coarse to cover the diverse reasoning patterns required by different tasks.
As the number of skills increases from 3 to 6 and finally to 9, performance gradually recovers and improves, indicating that DABStep benefits from a sufficiently diverse and fine-grained skill set.
These results demonstrate that effective skill discovery requires not only extracting useful procedures, but also maintaining an appropriate level of granularity to balance generalization and task-specific specialization.

\subsection{Data-Analytic Agent Analysis}
We examine whether the discovered skills are tied to a specific data-analytic agent.
To this end, we use GPT-5.2 and DeepSeek-V4-Pro as data-analytic agents to produce exploration trajectories, and then evaluate the resulting skills on different models.
As shown in Table~\ref{fig:data_agent_analysis}, \ours{} consistently improves over the Skill Creator baseline across all four combinations.
When GPT-5.2 is used for trajectories generation, the discovered skill improves Qwen3.5-397B and GPT-5.2 by 5.33\% and 7.69\% respectively.
When DeepSeek is used as the data-analytic agent, the gains further increase to 8.88\% and 10.06\%.
These results indicates that \ours{} is not specific to a particular model as a data-analytic agent, but has general applicability.
We also observe that using DeepSeek as the data-analytic agent leads to stronger downstream performance, indicating that higher-quality exploration trajectories may provide more informative evidence for skill discovery.

\subsection{Supervised Skill Discovery Analysis}
We study whether labeled trajectories are necessary for skill discovery.
Here, \textit{Ours} denotes \ours{}, while the other settings add different numbers of randomly selected supervised trajectories to the Skill Creator method.
On the 10-K reporting benchmark, \ours{} outperforms all supervised baseline variants.
This suggests that, for report-style data analysis, a small amount of trajectory-level supervision may provide insufficient task-level feedback. For full supervision, the Skill Manager may overfit by extracting skills from only a subset of supervised trajectories. In contrast, our alternating optimization between the Data-Analytic Agent and the Checklist Agent supplies richer verification signals and alleviates trajectory-level overfitting.

On DABStep, \ours{} obtains 62.82\% without any labeled trajectory, substantially outperforming the one trajectory supervised baseline and achieving performance comparable to the two and three trajectories supervised baselines.
When all exploration trajectories are supervised, the baseline reaches 72.19\%, indicating that full supervision remains beneficial for fixed-answer reasoning tasks.
Nevertheless, \ours{} achieves competitive performance under zero annotation cost, demonstrating strong label efficiency.

\subsection{Cost Analysis}
We further examine the cost-effectiveness of the discovered skill under a fixed interaction budget. 
For a controlled comparison, both Claude Code and ReAct agents are capped at 15 interaction turns. 
As shown in Table~\ref{tab:skill_efficiency}, the discovered skill consistently reduces token consumption while improving task accuracy for both agent scaffolds. 
In no-skill setting, Qwen3.5-397B with ReAct consumes substantially fewer tokens than Claude Code, indicating that it provides a more token-efficient alternative for data-analytic exploration. 
After incorporating the discovered skill, the two agents exhibit nearly identical token consumption, while their accuracies become much closer. 
These results suggest that the skill offers reusable procedural guidance that suppresses redundant exploration and improves efficiency across different agent scaffolds.
\section{Related Work}
\label{app:related}
\subsection{Data-Analytic Agents.}
Data analysis agents are designed to autonomously execute end-to-end data analysis tasks~\cite{ddr, dsgym, dabstep, kramabench, dataagent_survey, ads_pcs}. To handle complex real-world scenarios, existing approaches can be broadly categorized into two paradigms. \textit{(i) Predefined Workflows:} This line of work primarily leverages the reasoning and coding capabilities of general-purpose Large Language Models (LLMs) to navigate structured analytical pipelines. Applications include data visualization~\cite{matplotagent}, insight and report generation~\cite{dagent, agentada, insightpilot, datastorm}, heterogeneous data analysis~\cite{datacopilot, agenticdata, dsstar, datacross}, Text-to-SQL~\cite{deepeyes} and general data science workflows~\cite{datainterpreter, datawiseagent}. \textit{(ii) Agentic Training:} Diverging from off-the-shelf models, this paradigm tailors specialized agents for data analysis~\cite{deepanalyze, datamind, dataanalysis-study}. Such approaches rely on curating high-quality datasets for supervised fine-tuning or reinforcement learning to internalize domain expertise.

Distinct from both rigid predefined workflows and resource-intensive model training, \ours{} focuses on generating \textit{reusable skills}. This approach enhances the underlying model's data analysis capabilities without tying it to specific pipelines or requiring costly domain-specific training.

\subsection{LLM Agent Skills.}
Modular and reusable skills distilled from real-world scenarios or trajectories have been shown to enhance agents' ability to solve similar tasks~\cite{WelcomeExp, agentskill_survey, sok_skill, agent_skills}. 
Prior work has explored skills in heterogeneous forms~\cite{skillx, pro_skill, rl_skill_lib, memskill, memp, reasoning_skills, skillrl}. 
More recently, structured skill paradigms such as Anthropic's Agent Skills~\cite{anthropic2026skills} represent skills as reusable multi-file documents with dynamic loading and tool compatibility, further motivating the study of skill construction, optimization, and management.
Existing studies can be broadly viewed from several complementary perspectives. 
One line of work focuses on skill induction and evolution, where reusable skills are automatically constructed or refined from execution traces, failure cases, interaction feedback, task contexts, or other behavioral signals~\cite{trace2skill, coevoskills, evoskill, skillclaw, skillforge, skillopt, memento_skills, autoskill, skillos, ctx2skill}. 
Another line extends skill representations beyond purely textual or procedural forms, for example by grounding procedural knowledge in multimodal state evidence for visual-agent decision making~\cite{mmskills}. 
A further line studies skill-library management, including skill organization, retrieval, routing, governance, and multi-skill orchestration at scale~\cite{skillsvote, skillnet, skillrouter, agentskillos}. 
While these works have advanced reusable skill construction and deployment across diverse agent settings, we focus on the data-analysis domain and study unsupervised skill discovery for data-analytic agents. 
\section{Conclusion}
In this work, we introduce \textbf{\ours{}}, an unsupervised verifier-guided framework for discovering reusable data-analysis skills from unlabeled exploration trajectories.
\ours{} coordinates a Data-Analytic Agent for task exploration, an Unsupervised Verifier for extracting unsupervised signals, and a Skill Manager for distilling skills from contrastive trajectory groups.
We instantiate an Adaptive Checklist Verifier with checklist-based refinement for report-style tasks and an Answer Agreement Verifier with self-consistency for reasoning-style tasks.
Experiments on Deep Data Research and DABStep show consistent held-out improvements.
Overall, \ours{} establishes unsupervised verifier-guided skill discovery as an effective paradigm for the autonomous improvement of data-analytic agents.








\bibliographystyle{IEEEtran}
\bibliography{custom}

\begin{thebibliography}{10}
\providecommand{\url}[1]{#1}
\csname url@samestyle\endcsname
\providecommand{\newblock}{\relax}
\providecommand{\bibinfo}[2]{#2}
\providecommand{\BIBentrySTDinterwordspacing}{\spaceskip=0pt\relax}
\providecommand{\BIBentryALTinterwordstretchfactor}{4}
\providecommand{\BIBentryALTinterwordspacing}{\spaceskip=\fontdimen2\font plus
\BIBentryALTinterwordstretchfactor\fontdimen3\font minus \fontdimen4\font\relax}
\providecommand{\BIBforeignlanguage}[2]{{%
\expandafter\ifx\csname l@#1\endcsname\relax
\typeout{** WARNING: IEEEtran.bst: No hyphenation pattern has been}%
\typeout{** loaded for the language `#1'. Using the pattern for}%
\typeout{** the default language instead.}%
\else
\language=\csname l@#1\endcsname
\fi
#2}}
\providecommand{\BIBdecl}{\relax}
\BIBdecl

\bibitem{dataagent_survey}
\BIBentryALTinterwordspacing
Y.~Zhu, L.~Wang, C.~Yang, X.~Lin, B.~Li, W.~Zhou, X.~Liu, Z.~Peng, T.~Luo, Y.~Li, C.~Chai, C.~Chen, S.~Di, J.~Fan, J.~Sun, N.~Tang, F.~Tsung, J.~Wang, C.~Wu, Y.~Xu, S.~Zhang, Y.~Zhang, X.~Zhou, G.~Li, and Y.~Luo, ``A survey of data agents: Emerging paradigm or overstated hype?'' \emph{CoRR}, vol. abs/2510.23587, 2025. [Online]. Available: \url{https://doi.org/10.48550/arXiv.2510.23587}
\BIBentrySTDinterwordspacing

\bibitem{dsagent_survey}
\BIBentryALTinterwordspacing
P.~Wang, Y.~Yu, K.~Chen, X.~Zhan, and H.~Wang, ``Large language model-based data science agent: {A} survey,'' \emph{CoRR}, vol. abs/2508.02744, 2025. [Online]. Available: \url{https://doi.org/10.48550/arXiv.2508.02744}
\BIBentrySTDinterwordspacing

\bibitem{dr_survey}
\BIBentryALTinterwordspacing
W.~Zhang, X.~Li, Y.~Zhang, P.~Jia, Y.~Wang, H.~Guo, Y.~Liu, and X.~Zhao, ``Deep research: {A} survey of autonomous research agents,'' \emph{CoRR}, vol. abs/2508.12752, 2025. [Online]. Available: \url{https://doi.org/10.48550/arXiv.2508.12752}
\BIBentrySTDinterwordspacing

\bibitem{datainterpreter}
\BIBentryALTinterwordspacing
S.~Hong, Y.~Lin, B.~Liu, B.~Liu, B.~Wu, C.~Zhang, D.~Li, J.~Chen, J.~Zhang, J.~Wang, L.~Zhang, L.~Zhang, M.~Yang, M.~Zhuge, T.~Guo, T.~Zhou, W.~Tao, R.~Tang, X.~Lu, X.~Zheng, X.~Liang, Y.~Fei, Y.~Cheng, Y.~Ni, Z.~Gou, Z.~Xu, Y.~Luo, and C.~Wu, ``Data interpreter: An {LLM} agent for data science,'' in \emph{Findings of the Association for Computational Linguistics, {ACL} 2025, Vienna, Austria, July 27 - August 1, 2025}, W.~Che, J.~Nabende, E.~Shutova, and M.~T. Pilehvar, Eds.\hskip 1em plus 0.5em minus 0.4em\relax Association for Computational Linguistics, 2025, pp. 19\,796--19\,821. [Online]. Available: \url{https://aclanthology.org/2025.findings-acl.1016/}
\BIBentrySTDinterwordspacing

\bibitem{agenticdata}
\BIBentryALTinterwordspacing
J.~Sun, G.~Li, P.~Zhou, Y.~Ma, J.~Xu, and Y.~Li, ``Agenticdata: An agentic data analytics system for heterogeneous data,'' \emph{CoRR}, vol. abs/2508.05002, 2025. [Online]. Available: \url{https://doi.org/10.48550/arXiv.2508.05002}
\BIBentrySTDinterwordspacing

\bibitem{dsstar}
\BIBentryALTinterwordspacing
J.~Nam, J.~Yoon, J.~Chen, and T.~Pfister, ``{DS-STAR:} data science agent via iterative planning and verification,'' \emph{CoRR}, vol. abs/2509.21825, 2025. [Online]. Available: \url{https://doi.org/10.48550/arXiv.2509.21825}
\BIBentrySTDinterwordspacing

\bibitem{agentada}
\BIBentryALTinterwordspacing
A.~Abaskohi, A.~V. Ramesh, S.~Nanisetty, C.~Goel, D.~V{\'{a}}zquez, C.~Pal, S.~Gella, G.~Carenini, and I.~H. Laradji, ``Agentada: Skill-adaptive data analytics for tailored insight discovery,'' \emph{CoRR}, vol. abs/2504.07421, 2025. [Online]. Available: \url{https://doi.org/10.48550/arXiv.2504.07421}
\BIBentrySTDinterwordspacing

\bibitem{datawiseagent}
\BIBentryALTinterwordspacing
Z.~You, Y.~Zhang, D.~Xu, Y.~Lou, Y.~Yan, W.~Wang, H.~Zhang, and Y.~Huang, ``Datawiseagent: {A} notebook-centric {LLM} agent framework for automated data science,'' \emph{CoRR}, vol. abs/2503.07044, 2025. [Online]. Available: \url{https://doi.org/10.48550/arXiv.2503.07044}
\BIBentrySTDinterwordspacing

\bibitem{datacopilot}
\BIBentryALTinterwordspacing
W.~Zhang, Y.~Shen, W.~Lu, and Y.~Zhuang, ``Data-copilot: Bridging billions of data and humans with autonomous workflow,'' \emph{CoRR}, vol. abs/2306.07209, 2023. [Online]. Available: \url{https://doi.org/10.48550/arXiv.2306.07209}
\BIBentrySTDinterwordspacing

\bibitem{dagent}
\BIBentryALTinterwordspacing
W.~Xu, Y.~Mao, X.~Zhang, C.~Zhang, X.~Dong, M.~Zhang, and Y.~Gao, ``Dagent: {A} relational database-driven data analysis report generation agent,'' \emph{CoRR}, vol. abs/2503.13269, 2025. [Online]. Available: \url{https://doi.org/10.48550/arXiv.2503.13269}
\BIBentrySTDinterwordspacing

\bibitem{datamind}
\BIBentryALTinterwordspacing
S.~Qiao, Y.~Zhao, Z.~Qiu, X.~Wang, J.~Zhang, Z.~Bin, N.~Zhang, Y.~Jiang, P.~Xie, F.~Huang, and H.~Chen, ``Scaling generalist data-analytic agents,'' \emph{CoRR}, vol. abs/2509.25084, 2025. [Online]. Available: \url{https://doi.org/10.48550/arXiv.2509.25084}
\BIBentrySTDinterwordspacing

\bibitem{deepanalyze}
\BIBentryALTinterwordspacing
S.~Zhang, J.~Fan, M.~Fan, G.~Li, and X.~Du, ``Deepanalyze: Agentic large language models for autonomous data science,'' \emph{CoRR}, vol. abs/2510.16872, 2025. [Online]. Available: \url{https://doi.org/10.48550/arXiv.2510.16872}
\BIBentrySTDinterwordspacing

\bibitem{anthropic2026skills}
\BIBentryALTinterwordspacing
Anthropic, ``What are skills?'' Claude Help Center, 2026, accessed: 2026-05-03. [Online]. Available: \url{https://support.claude.com/en/articles/12512176-what-are-skills}
\BIBentrySTDinterwordspacing

\bibitem{sok_skill}
\BIBentryALTinterwordspacing
Y.~Jiang, D.~Li, H.~Deng, B.~Ma, X.~Wang, Q.~Wang, and G.~Yu, ``Sok: Agentic skills - beyond tool use in {LLM} agents,'' \emph{CoRR}, vol. abs/2602.20867, 2026. [Online]. Available: \url{https://doi.org/10.48550/arXiv.2602.20867}
\BIBentrySTDinterwordspacing

\bibitem{agent_skills}
\BIBentryALTinterwordspacing
G.~F. Ling, S.~Zhong, and R.~L. Huang, ``Agent skills: A data-driven analysis of claude skills for extending large language model functionality,'' \emph{ArXiv}, vol. abs/2602.08004, 2026. [Online]. Available: \url{https://api.semanticscholar.org/CorpusID:285453033}
\BIBentrySTDinterwordspacing

\bibitem{skillx}
\BIBentryALTinterwordspacing
C.~Wang, Z.~Yu, X.~Xie, W.~Yao, R.~Fang, S.~Qiao, K.~Cao, G.~Zheng, X.~Qi, P.~Zhang, and S.~Deng, ``Skillx: Automatically constructing skill knowledge bases for agents,'' 2026. [Online]. Available: \url{https://api.semanticscholar.org/CorpusID:287204111}
\BIBentrySTDinterwordspacing

\bibitem{trace2skill}
\BIBentryALTinterwordspacing
J.~Ni, Y.~Liu, X.~Liu, Y.~Sun, M.~Zhou, P.~Cheng, D.~Wang, E.~Zhao, X.~Jiang, and G.~Jiang, ``Trace2skill: Distill trajectory-local lessons into transferable agent skills,'' \emph{CoRR}, vol. abs/2603.25158, 2026. [Online]. Available: \url{https://doi.org/10.48550/arXiv.2603.25158}
\BIBentrySTDinterwordspacing

\bibitem{evoskill}
\BIBentryALTinterwordspacing
S.~Alzubi, N.~Provenzano, J.~Bingham, W.~Chen, and T.~Vu, ``Evoskill: Automated skill discovery for multi-agent systems,'' \emph{CoRR}, vol. abs/2603.02766, 2026. [Online]. Available: \url{https://doi.org/10.48550/arXiv.2603.02766}
\BIBentrySTDinterwordspacing

\bibitem{coevoskills}
\BIBentryALTinterwordspacing
H.~Zhang, S.~Fan, H.~P. Zou, Y.~Chen, Z.~Wang, J.~Zhou, C.~Li, W.-C. Huang, Y.~Yao, K.~Zheng, X.~Liu, X.~Li, and P.~S. Yu, ``Coevoskills: Self-evolving agent skills via co-evolutionary verification,'' 2026. [Online]. Available: \url{https://api.semanticscholar.org/CorpusID:287071917}
\BIBentrySTDinterwordspacing

\bibitem{skillopt}
\BIBentryALTinterwordspacing
Y.~Yang, Z.~Gong, W.~Huang, Q.~Yang, Z.~Zhou, Z.~Huang, Y.~Li, X.~Gao, Q.~Dai, B.~Liu, K.~Qiu, Y.~Yang, D.~Chen, X.-T. Yang, and C.~Luo, ``Skillopt: Executive strategy for self-evolving agent skills,'' 2026. [Online]. Available: \url{https://api.semanticscholar.org/CorpusID:288652900}
\BIBentrySTDinterwordspacing

\bibitem{skillclaw}
\BIBentryALTinterwordspacing
Z.~Ma, S.~Yang, Y.~Ji, X.~Wang, Y.~Wang, Y.~Hu, T.~Huang, and X.~Chu, ``Skillclaw: Let skills evolve collectively with agentic evolver,'' 2026. [Online]. Available: \url{https://api.semanticscholar.org/CorpusID:287256390}
\BIBentrySTDinterwordspacing

\bibitem{skillos}
\BIBentryALTinterwordspacing
S.~Ouyang, J.~Yan, Y.~Chen, R.~Han, Z.~Wang, B.~Dalvi, R.~Meng, C.-L. Li, Y.~Jiao, K.~Zha, M.~Shen, V.~Tirumalashetty, G.~Lee, J.~Han, T.~Pfister, and C.-Y. Lee, ``Skillos: Learning skill curation for self-evolving agents,'' 2026. [Online]. Available: \url{https://api.semanticscholar.org/CorpusID:288014414}
\BIBentrySTDinterwordspacing

\bibitem{ddr}
\BIBentryALTinterwordspacing
W.~Liu, P.~Yu, M.~Orini, Y.~Du, and Y.~He, ``Hunt instead of wait: Evaluating deep data research on large language models,'' \emph{CoRR}, vol. abs/2602.02039, 2026. [Online]. Available: \url{https://doi.org/10.48550/arXiv.2602.02039}
\BIBentrySTDinterwordspacing

\bibitem{dabstep}
\BIBentryALTinterwordspacing
A.~Egg, M.~I. Goyanes, F.~Kingma, A.~Mora, L.~von Werra, and T.~Wolf, ``Dabstep: Data agent benchmark for multi-step reasoning,'' \emph{CoRR}, vol. abs/2506.23719, 2025. [Online]. Available: \url{https://doi.org/10.48550/arXiv.2506.23719}
\BIBentrySTDinterwordspacing

\bibitem{react}
\BIBentryALTinterwordspacing
S.~Yao, J.~Zhao, D.~Yu, N.~Du, I.~Shafran, K.~R. Narasimhan, and Y.~Cao, ``React: Synergizing reasoning and acting in language models,'' in \emph{The Eleventh International Conference on Learning Representations, {ICLR} 2023, Kigali, Rwanda, May 1-5, 2023}.\hskip 1em plus 0.5em minus 0.4em\relax OpenReview.net, 2023. [Online]. Available: \url{https://openreview.net/forum?id=WE\_vluYUL-X}
\BIBentrySTDinterwordspacing

\bibitem{gpt-5}
\BIBentryALTinterwordspacing
OpenAI, ``Openai {GPT-5} system card,'' \emph{CoRR}, vol. abs/2601.03267, 2026. [Online]. Available: \url{https://doi.org/10.48550/arXiv.2601.03267}
\BIBentrySTDinterwordspacing

\bibitem{anthropic2026skillcreator}
\BIBentryALTinterwordspacing
Anthropic, ``{Skill Creator},'' 2026, accessed: 2026-06-03. [Online]. Available: \url{https://github.com/anthropics/skills/blob/main/skills/skill-creator/SKILL.md}
\BIBentrySTDinterwordspacing

\bibitem{anthropic2026claudesonnet46systemcard}
\BIBentryALTinterwordspacing
------, ``{System Card: Claude Sonnet 4.6},'' Anthropic Model System Cards, 2026, accessed: 2026-06-03. [Online]. Available: \url{https://www-cdn.anthropic.com/bbd8ef16d70b7a1665f14f306ee88b53f686aa75.pdf}
\BIBentrySTDinterwordspacing

\bibitem{anthropic2025claudesonnet45systemcard}
\BIBentryALTinterwordspacing
------, ``{System Card: Claude Sonnet 4.5},'' Anthropic Model System Cards, 2025, accessed: 2026-05-17. [Online]. Available: \url{https://www-cdn.anthropic.com/963373e433e489a87a10c823c52a0a013e9172dd.pdf}
\BIBentrySTDinterwordspacing

\bibitem{deepseek-v4}
DeepSeek-AI, ``Deepseek-v4: Towards highly efficient million-token context intelligence,'' 2026.

\bibitem{qwen35}
\BIBentryALTinterwordspacing
Q.~Team, ``Qwen3.5: Accelerating productivity with native multimodal agents,'' February 2026. [Online]. Available: \url{https://qwen.ai/blog?id=qwen3.5}
\BIBentrySTDinterwordspacing

\bibitem{dsgym}
\BIBentryALTinterwordspacing
F.~Nie, J.~Wang, H.~Hua, F.~Bianchi, Y.~Kwon, Z.~Qi, O.~Queen, S.~Zhu, and J.~Zou, ``Dsgym: {A} holistic framework for evaluating and training data science agents,'' \emph{CoRR}, vol. abs/2601.16344, 2026. [Online]. Available: \url{https://doi.org/10.48550/arXiv.2601.16344}
\BIBentrySTDinterwordspacing

\bibitem{kramabench}
\BIBentryALTinterwordspacing
E.~Lai, G.~Vitagliano, Z.~Zhang, S.~Sudhir, O.~Chabra, A.~Zeng, A.~A. Zabreyko, C.~Li, F.~Kossmann, J.~Ding, J.~Chen, M.~Markakis, M.~Russo, W.~Wang, Z.~Wu, M.~J. Cafarella, L.~Cao, S.~Madden, and T.~Kraska, ``Kramabench: {A} benchmark for {AI} systems on data-to-insight pipelines over data lakes,'' \emph{CoRR}, vol. abs/2506.06541, 2025. [Online]. Available: \url{https://doi.org/10.48550/arXiv.2506.06541}
\BIBentrySTDinterwordspacing

\bibitem{ads_pcs}
\BIBentryALTinterwordspacing
Z.~T. Rewolinski, A.~Zane, H.~Huang, C.~Singh, C.~Wang, J.~Gao, and B.~Yu, ``Sanity checks for agentic data science,'' 2026. [Online]. Available: \url{https://api.semanticscholar.org/CorpusID:287433410}
\BIBentrySTDinterwordspacing

\bibitem{matplotagent}
\BIBentryALTinterwordspacing
Z.~Yang, Z.~Zhou, S.~Wang, X.~Cong, X.~Han, Y.~Yan, Z.~Liu, Z.~Tan, P.~Liu, D.~Yu, Z.~Liu, X.~Shi, and M.~Sun, ``Matplotagent: Method and evaluation for llm-based agentic scientific data visualization,'' in \emph{Findings of the Association for Computational Linguistics, {ACL} 2024, Bangkok, Thailand and virtual meeting, August 11-16, 2024}, L.~Ku, A.~Martins, and V.~Srikumar, Eds.\hskip 1em plus 0.5em minus 0.4em\relax Association for Computational Linguistics, 2024, pp. 11\,789--11\,804. [Online]. Available: \url{https://doi.org/10.18653/v1/2024.findings-acl.701}
\BIBentrySTDinterwordspacing

\bibitem{insightpilot}
\BIBentryALTinterwordspacing
P.~Ma, R.~Ding, S.~Wang, S.~Han, and D.~Zhang, ``Insightpilot: An llm-empowered automated data exploration system,'' in \emph{Proceedings of the 2023 Conference on Empirical Methods in Natural Language Processing, {EMNLP} 2023 - System Demonstrations, Singapore, December 6-10, 2023}, Y.~Feng and E.~Lefever, Eds.\hskip 1em plus 0.5em minus 0.4em\relax Association for Computational Linguistics, 2023, pp. 346--352. [Online]. Available: \url{https://doi.org/10.18653/v1/2023.emnlp-demo.31}
\BIBentrySTDinterwordspacing

\bibitem{datastorm}
\BIBentryALTinterwordspacing
S.~Liu, Y.~Jiang, S.~Farook, C.~N. Sanchez, D.~F.~C. Pena, and M.~S. Lam, ``Datastorm: Deep research on large-scale databases using exploratory data analysis and data storytelling,'' 2026. [Online]. Available: \url{https://api.semanticscholar.org/CorpusID:287248168}
\BIBentrySTDinterwordspacing

\bibitem{datacross}
\BIBentryALTinterwordspacing
R.~Qi, Z.~Liu, and W.~Zhang, ``Datacross: A unified benchmark and agent framework for cross-modal heterogeneous data analysis,'' \emph{ArXiv}, vol. abs/2601.21403, 2026. [Online]. Available: \url{https://api.semanticscholar.org/CorpusID:285140426}
\BIBentrySTDinterwordspacing

\bibitem{deepeyes}
\BIBentryALTinterwordspacing
B.~Li, C.~Chen, Z.~Xue, Y.~Mei, and Y.~Luo, ``Deepeye-sql: {A} software-engineering-inspired text-to-sql framework,'' \emph{CoRR}, vol. abs/2510.17586, 2025. [Online]. Available: \url{https://doi.org/10.48550/arXiv.2510.17586}
\BIBentrySTDinterwordspacing

\bibitem{dataanalysis-study}
\BIBentryALTinterwordspacing
Y.~Zhu, Y.~Zhong, J.~Zhang, Z.~Zhang, S.~Qiao, Y.~Luo, L.~Du, D.~Zheng, H.~Chen, and N.~Zhang, ``Why do open-source llms struggle with data analysis? {A} systematic empirical study,'' \emph{CoRR}, vol. abs/2506.19794, 2025. [Online]. Available: \url{https://doi.org/10.48550/arXiv.2506.19794}
\BIBentrySTDinterwordspacing

\bibitem{WelcomeExp}
\BIBentryALTinterwordspacing
D.~Silver and R.~Sutton, ``Welcome to the era of experience.'' [Online]. Available: \url{https://api.semanticscholar.org/CorpusID:277919528}
\BIBentrySTDinterwordspacing

\bibitem{agentskill_survey}
\BIBentryALTinterwordspacing
R.~Xu and Y.~Yan, ``Agent skills for large language models: Architecture, acquisition, security, and the path forward,'' \emph{CoRR}, vol. abs/2602.12430, 2026. [Online]. Available: \url{https://doi.org/10.48550/arXiv.2602.12430}
\BIBentrySTDinterwordspacing

\bibitem{pro_skill}
\BIBentryALTinterwordspacing
Z.~Z. Wang, A.~Gandhi, G.~Neubig, and D.~Fried, ``Inducing programmatic skills for agentic tasks,'' \emph{CoRR}, vol. abs/2504.06821, 2025. [Online]. Available: \url{https://doi.org/10.48550/arXiv.2504.06821}
\BIBentrySTDinterwordspacing

\bibitem{rl_skill_lib}
\BIBentryALTinterwordspacing
J.~Wang, Q.~Yan, Y.~Wang, Y.~Tian, S.~S. Mishra, Z.~Xu, M.~Gandhi, P.~Xu, and L.~L. Cheong, ``Reinforcement learning for self-improving agent with skill library,'' \emph{CoRR}, vol. abs/2512.17102, 2025. [Online]. Available: \url{https://doi.org/10.48550/arXiv.2512.17102}
\BIBentrySTDinterwordspacing

\bibitem{memskill}
\BIBentryALTinterwordspacing
H.~Zhang, Q.~Long, J.~Bao, T.~Feng, W.~Zhang, H.~Yue, and W.~Wang, ``Memskill: Learning and evolving memory skills for self-evolving agents,'' \emph{CoRR}, vol. abs/2602.02474, 2026. [Online]. Available: \url{https://doi.org/10.48550/arXiv.2602.02474}
\BIBentrySTDinterwordspacing

\bibitem{memp}
\BIBentryALTinterwordspacing
R.~Fang, Y.~Liang, X.~Wang, J.~Wu, S.~Qiao, P.~Xie, F.~Huang, H.~Chen, and N.~Zhang, ``Memp: Exploring agent procedural memory,'' \emph{CoRR}, vol. abs/2508.06433, 2025. [Online]. Available: \url{https://doi.org/10.48550/arXiv.2508.06433}
\BIBentrySTDinterwordspacing

\bibitem{reasoning_skills}
\BIBentryALTinterwordspacing
G.~Zhao, Q.~Shi, X.~Xiao, X.~Zhang, T.~Yang, and L.~Sun, ``Thinking with reasoning skills: Fewer tokens, more accuracy,'' \emph{CoRR}, vol. abs/2604.21764, 2026. [Online]. Available: \url{https://doi.org/10.48550/arXiv.2604.21764}
\BIBentrySTDinterwordspacing

\bibitem{skillrl}
\BIBentryALTinterwordspacing
P.~Xia, J.~Chen, H.~Wang, J.~Liu, K.~Zeng, Y.~Wang, S.~Han, Y.~Zhou, X.~Zhao, H.~Chen, Z.~Zheng, C.~Xie, and H.~Yao, ``Skillrl: Evolving agents via recursive skill-augmented reinforcement learning,'' \emph{ArXiv}, vol. abs/2602.08234, 2026. [Online]. Available: \url{https://api.semanticscholar.org/CorpusID:285452037}
\BIBentrySTDinterwordspacing

\bibitem{skillforge}
\BIBentryALTinterwordspacing
X.~Liu, X.~Luo, L.~Li, G.~Huang, J.~Liu, and H.~Qiao, ``Skillforge: Forging domain-specific, self-evolving agent skills in cloud technical support,'' 2026. [Online]. Available: \url{https://api.semanticscholar.org/CorpusID:287351631}
\BIBentrySTDinterwordspacing

\bibitem{memento_skills}
\BIBentryALTinterwordspacing
H.~Zhou, S.~Guo, A.~Liu, Z.~Yu, Z.~Gong, B.~Zhao, Z.~Chen, M.~Zhang, Y.~Chen, J.~Li, R.~Yang, Q.~Liu, X.~Yu, J.~Zhou, N.~Wang, C.~Sun, and J.~Wang, ``Memento-skills: Let agents design agents,'' 2026. [Online]. Available: \url{https://api.semanticscholar.org/CorpusID:286673350}
\BIBentrySTDinterwordspacing

\bibitem{autoskill}
\BIBentryALTinterwordspacing
Y.~Yang, J.~Li, Q.~Pan, B.~Zhan, Y.~Cai, L.~Du, J.~Zhou, K.~Chen, Q.~Chen, X.~Li, B.~Zhang, and L.~He, ``Autoskill: Experience-driven lifelong learning via skill self-evolution,'' \emph{ArXiv}, vol. abs/2603.01145, 2026. [Online]. Available: \url{https://api.semanticscholar.org/CorpusID:286224498}
\BIBentrySTDinterwordspacing

\bibitem{ctx2skill}
\BIBentryALTinterwordspacing
S.~Si, H.~Zhao, Y.~Lei, Q.~Wang, D.~Chen, Z.~Wang, Z.~Wang, K.~Luo, Z.~Wang, G.~Chen, F.~Qi, M.~Zhang, and M.~Sun, ``From context to skills: Can language models learn from context skillfully?'' 2026. [Online]. Available: \url{https://api.semanticscholar.org/CorpusID:287915777}
\BIBentrySTDinterwordspacing

\bibitem{mmskills}
\BIBentryALTinterwordspacing
K.~Zhang, S.~Shao, Q.~Li, J.~Lin, L.~Fu, S.~Wang, W.~Jiao, Y.~Lu, W.~Liu, W.~Zhang, and Y.~Yu, ``Mmskills: Towards multimodal skills for general visual agents,'' 2026. [Online]. Available: \url{https://api.semanticscholar.org/CorpusID:288254572}
\BIBentrySTDinterwordspacing

\bibitem{skillsvote}
\BIBentryALTinterwordspacing
H.~Liu, H.~Yang, T.~Jiang, B.~Tang, F.~Xiong, and Z.~Li, ``Skillsvote: Lifecycle governance of agent skills from collection, recommendation to evolution,'' 2026. [Online]. Available: \url{https://api.semanticscholar.org/CorpusID:288651284}
\BIBentrySTDinterwordspacing

\bibitem{skillnet}
\BIBentryALTinterwordspacing
Y.~Liang, R.~Zhong, H.~Xu, C.~Jiang, Y.~Zhong, R.~Fang, J.~Gu, S.~Deng, Y.~Yao, M.~Wang, S.~Qiao, X.~Xu, T.~Wu, K.~Wang, Y.~Liu, Z.~Bi, J.~Lou, Y.~E. Jiang, H.~Zhu, G.~Yu, H.~Hong, L.~Huang, H.~Xue, C.~Wang, Y.~Wang, Z.~Shan, X.~Chen, Z.~Tu, F.~Xiong, X.~Xie, P.~Zhang, Z.~Gui, L.~Liang, J.~Zhou, C.~Wu, J.~Shang, Y.~Gong, J.~Lin, C.~Xu, H.~Deng, W.~Zhang, K.~Ding, Q.~Zhang, F.~Huang, N.~Zhang, J.~Z. Pan, G.~Qi, H.~Wang, and H.~Chen, ``Skillnet: Create, evaluate, and connect {AI} skills,'' \emph{CoRR}, vol. abs/2603.04448, 2026. [Online]. Available: \url{https://doi.org/10.48550/arXiv.2603.04448}
\BIBentrySTDinterwordspacing

\bibitem{skillrouter}
\BIBentryALTinterwordspacing
Y.~Zheng, Z.~Zhang, C.~Ma, Y.~Yu, J.~Zhu, Y.~Wu, T.~Xu, B.~Dong, H.~Zhu, R.~Huang, and G.~Yu, ``Skillrouter: Skill routing for llm agents at scale,'' 2026. [Online]. Available: \url{https://api.semanticscholar.org/CorpusID:286770530}
\BIBentrySTDinterwordspacing

\bibitem{agentskillos}
\BIBentryALTinterwordspacing
H.~Li, C.~Mu, J.~Chen, S.~Ren, Z.~Cui, Y.~Zhang, L.~Bai, and S.~Hu, ``Organizing, orchestrating, and benchmarking agent skills at ecosystem scale,'' 2026. [Online]. Available: \url{https://api.semanticscholar.org/CorpusID:286222444}
\BIBentrySTDinterwordspacing

\end{thebibliography}

\end{document}